\documentclass[%
	final,
	notitlepage,
	narroweqnarray,
	inline,
	twoside,
   	]{ieee}

\usepackage{amsmath}
\usepackage{amssymb}
\usepackage{mathabx}
\usepackage{algorithmic}
\usepackage{algorithm}
\usepackage{subfigure}
\usepackage{graphicx}
\usepackage{url}
\usepackage{lscape}
\usepackage{epsfig}
\usepackage{subfigure}
\usepackage{lscape}
\usepackage{booktabs}
\usepackage{bm}
\usepackage{epstopdf}

\DeclareMathOperator{\sgn}{sgn}

\DeclareMathOperator{\tr}{tr}
\DeclareMathOperator{\diag}{diag}

\begin{document}
\graphicspath{{figs/}}
\title[Learning to Hash for Indexing Big Data - A Survey]
{Learning to Hash for Indexing Big Data - A Survey}
   
\author[WANG, LIU, KUMAR, AND CHANG]{
Jun Wang\member{Member},
       \authorinfo{Jun Wang is with the Institute of Data Science and Technology at Alibaba Group, Seattle, WA, 98101, USA.
       Phone: $+$1\,917\,945--2619, e-mail: j.wang@alibaba-inc.com.}
\and Wei Liu\member{Member},
       \authorinfo{Wei Liu is with IBM T. J. Watson Research Center, Yorktown Heights, NY 10598, USA.
       Phone: $+$1\,917\,945--1274, e-mail: weiliu@us.ibm.com.}
\and Sanjiv Kumar\member{Member},
       \authorinfo{Sanjiv Kumar is with Google Research, New York, NY 10011, USA.
       Phone: $+$1\,212\,865--2214, e-mail: sanjivk@google.com.}
\and and Shih-Fu Chang\member{Fellow}
       \authorinfo{Shih-Fu Chang is with the Departments of Electrical
       Engineering and Computer Science, Columbia University, New York, NY 10027, USA.
       Phone: $+$1\,212\,854--6894, fax: $+$1\,212\,932--9421, e-mail: sfchang@ee.columbia.edu.}
}

\journal{PROCEEDINGS OF THE IEEE}
\ieeecopyright{0018-9219/\$26.00 \copyright\ 2011 IEEE}
\lognumber{xxxxxxx}
\loginfo{Manuscript received Oct. XX, 2014.}
\firstpage{1}


\maketitle

\begin{abstract}
The explosive growth in big data has attracted much attention in designing efficient indexing and search methods recently. In many critical applications such as large-scale search and pattern matching, finding the nearest neighbors to a query is a fundamental research problem. However, the straightforward solution using exhaustive comparison is infeasible due to the prohibitive computational complexity and memory requirement. In response, Approximate Nearest Neighbor (ANN) search based on hashing techniques has become popular due to its promising performance in both efficiency and accuracy. Prior randomized hashing methods, e.g., Locality-Sensitive Hashing (LSH), explore data-independent hash functions with random projections or permutations. Although having elegant theoretic guarantees on the search quality in certain metric spaces, performance of randomized hashing has been shown insufficient in many real-world applications. As a remedy, new approaches incorporating data-driven learning methods in development of advanced hash functions have emerged. Such learning to hash methods exploit information such as data distributions or class labels when optimizing the hash codes or functions. Importantly, the learned hash codes are able to preserve the proximity of neighboring data in the original feature spaces in the hash code spaces. The goal of this paper is to provide readers with systematic understanding of insights, pros and cons of the emerging techniques. We provide a comprehensive survey of the learning to hash framework and representative techniques of various types, including unsupervised, semi-supervised, and supervised. In addition, we also summarize recent hashing approaches utilizing the deep learning models. Finally, we discuss the future direction and trends of research in this area.
\end{abstract}


\section{Introduction}
The advent of Internet has resulted in massive information overloading in the recent decades. Nowadays, the World Wide Web has over $366$ million accessible websites, containing more than 1 trillion webpages\footnote{The number webpages is estimated based on the number of indexed links by Google in 2008.}. For instance, {\it Twitter} receives over $100$ million tweets per day, and {\cal Yahoo!} exchanges over $3$ billion messages per day. Besides the overwhelming textual data, the photo sharing website {\it Flickr} has more than $5$ billion images available, where images are still being uploaded at the rate of over $3,000$ images per minute. Another rich media sharing website {\it YouTube} receives more than $100$ hours of videos uploaded per minute. Due to the dramatic increase in the size of the data, modern information technology infrastructure has to deal with such gigantic databases. In fact, compared to the cost of storage, searching for relevant content in massive databases turns out to be even a more challenging task. In particular, searching for rich media data, such as audio, images, and videos, remains a major challenge since there exist major gaps between available solutions and practical needs in both accuracy and computational costs. Besides the widely used text-based commercial search engines such as {\it Google} and {\it Bing}, \textit{content-based image retrieval} (CBIR) has attracted substantial attention in the past decade~\cite{Datta:2008ACM}. Instead of relying on textual keywords based indexing structures, CBIR requires efficiently indexing media content in order to directly respond to visual queries.

Searching for similar data samples in a given database essentially relates to the fundamental problem of nearest neighbor search~\cite{Shakhnarovich:2006book}. Exhaustively comparing a query point $q$ with each sample in a database ${\cal X}$ is infeasible because the linear time complexity ${\cal O}(|{\cal X}|)$ tends to be expensive in realistic large-scale settings. Besides the scalability issue, most practical large-scale applications also suffer from the {\it curse of dimensionality}~\cite{Bellman:1957dynamic}, since data under modern analytics usually contains thousands or even tens of thousands of dimensions, e.g., in documents and images. Therefore, beyond the infeasibility of the computational cost for exhaustive search, the storage constraint originating from loading original data into memory also becomes a critical bottleneck. Note that retrieving a set of \textit{Approximate Nearest Neighbors} (ANNs) is often sufficient for many practical applications. Hence, a fast and effective indexing method achieving sublinear ($o(|{\cal X}|)$), logarithmic (${\cal O}(\log |{\cal X}|)$), or even constant (${\cal O}(1)$) query time is desired for ANN search. Tree-based indexing approaches, such as KD tree~\cite{Bentley:1975CACM}, ball tree~\cite{Omohundro:1987CS}, metric tree~\cite{Uhlmann:1991IPL}, and vantage point tree~\cite{Yianilos:1993SODA}, have been popular during the past several decades. However, tree-based approaches require significant storage costs (sometimes more than the data itself). In addition, the performance of tree-based indexing methods dramatically degrades when handling high-dimensional data~\cite{Goodman:2004HandBook}. More recently, product quantization techniques have been proposed to encode high-dimensional data vectors via subspace decomposition for efficient ANN search~\cite{Jegou2011product}\cite{OPQ:cvpr2013}.

Unlike the recursive partitioning used by tree-based indexing methods, hashing methods repeatedly partition the entire dataset and derive a single hash 'bit'\footnote{Depending on the type of the hash function used, each hash may return  either an integer or simply a binary bit. In this survey we primarily focus on binary hashing techniques as they are used most commonly due to their computational and storage efficiency.} from each partitioning. In binary partitioning based hashing, input data is mapped to a discrete code space called Hamming space, where each sample is represented by a  binary code. Specifically, given $N$ $D$-dim vectors ${\bf X}\in{\mathbb R}^{D\times N}$, the goal of hashing is to derive suitable $K$-bit binary codes ${\bf Y}\in{\mathbb B}^{K\times N}$. To generate ${\bf Y}$, $K$ binary hash functions $\big\{h_k:\mathbb{R}^D\mapsto\mathbb{B}\big\}_{k=1}^K$ are needed. Note that hashing-based ANN search techniques can lead to substantially reduced storage as they usually store only compact binary codes. For instance, $80$ million tiny images ($32\times 32$ pixels, double type) cost around $600$G bytes~\cite{Torralba:2008PAMI}, but can be compressed into $64$-bit binary codes requiring only $600$M bytes! In many cases, hash codes are organized into a hash table for inverse table lookup, as shown in Figure~\ref{fig:hashdigram}. One advantage of hashing-based indexing is that hash table lookup takes only constant query time. In fact, in many cases, another alternative way of finding the nearest neighbors in the code space by explicitly computing Hamming distance with all the database items can be done very efficiently as well.

Hashing methods have been intensively studied and widely used in many different fields, including computer graphics, computational geometry, telecommunication, computer vision, \textit{etc.}, for several decades~\cite{Knuth:1997art}. Among these methods, the randomized scheme of Locality-Sensitive Hashing (LSH) is one of the most popular choices~\cite{Gionis:1999VLDB}. A key ingredient in LSH family of techniques is a hash function that, with high probabilities, returns the same bit for the nearby data points in the original metric space. LSH provides interesting asymptotic theoretical properties leading to performance guarantees. However, LSH based randomized techniques suffer from several crucial drawbacks. First, to achieve desired search precision, LSH often needs to use long hash codes, which reduces the recall. Multiple hash tables are used to alleviate this issue, but it dramatically increases the storage cost as well as the query time. Second, the theoretical guarantees of LSH only apply to certain metrics such as $\ell_p$ ($p \in (0, 2]$) and Jaccard ~\cite{SSH:2012PAMI}. However, returning ANNs in such metric spaces may not lead to good search performance when semantic similarity is represented in a complex way instead of a simple distance or similarity metric. This discrepancy between semantic and metric spaces has been recognized in the computer vision and machine learning communities, namely as {\em semantic gap}~\cite{Smeulders:2000PAMI}.

To tackle the aforementioned issues, many hashing methods have been proposed recently to leverage machine learning techniques to produce more effective hash codes~\cite{Cayton:2007NIPS}. The goal of {\em learning to hash} is to learn data-dependent and task-specific hash functions that yield compact binary codes to achieve good search accuracy~\cite{CHJO:2011CVPR}. In order to achieve this goal, sophisticated machine learning tools and algorithms have been adapted to the procedure of hash function design, including the boosting algorithm~\cite{Shakhnarovich:2005learning}, distance metric learning~\cite{Kulis:2009PAMI}, asymmetric binary embedding~\cite{ADBE:2011CVPR}, kernel methods~\cite{Kulis:2011PAMI}\cite{SKH:cvpr2012}, compressed sensing~\cite{CompH:cvpr2013}, maximum margin learning~\cite{RMMH:2011CVPR}, sequential learning~\cite{SPLH:icml2010}, clustering analysis~\cite{KMH:cvpr2013}, semi-supervised learning~\cite{SSH:2012PAMI}, supervised learning~\cite{Kulis:2009NIPS}\cite{SKH:cvpr2012}, graph learning~\cite{GH:icml2011}, and so on. For instance, in the specific application of image search, the similarity (or distance) between image pairs is usually not defined via a simple metric. Ideally, one would like to provide pairs of images that contain ``similar'' or ``dissimilar'' images. From such pairwise labeled information, a good hashing mechanism should be able to generate hash codes which preserve the semantic consistency, \textit{i.e.}, semantically similar images should have similar codes. Both the supervised and semi-supervised learning paradigms have been explored using such pairwise semantic relationships to learn semantically relevant hash functions~\cite{Torralba:2008PAMI}\cite{SSH:cvpr2010}\cite{Norouzi:2011ICML}\cite{Xu:2011ICCV}. In this paper, we will survey important representative hashing approaches and also discuss the future research directions.

The remainder of this article is organized as follows. In Section~\ref{sec:background}, we will present necessary background information, prior randomized hashing methods, and the motivations of studying hashing. Section~\ref{sec:overview} gives a high-level overview of emerging learning-based hashing methods. In Section~\ref{sec:methods}, we survey several popular methods that fall into the {\em learning to hash} framework. In addition, Section~\ref{sec:dnnhash} describes the recent development of using neural networks to perform deep learning of hash codes. Section~\ref{sec:advances} discusses advanced hashing techniques and large-scale applications of hashing. Several  open issues and future directions are described in Section~\ref{sec:conclusion}.

\section{Notations and Background}
\label{sec:background}
In this section, we will first present the notations, as summarized in Table~\ref{Tab:notations}. Then we will briefly introduce the conceptual paradigm of hashing-based ANN search. Finally, we will present some background information on hashing methods, including the introduction of two well-known randomized hashing techniques. 

\begin{table*}[t]
\setlength{\tabcolsep}{4pt} \centering
\caption{Summary Of Notations}
\begin{tabular}{lr}
\toprule
Symbol & Definition\\
\midrule
$N$& number of data points\\
$D$& dimensionality of data points\\
$K$& number of hash bits\\
$i, j$& indices of data points\\
$k$& index of a hash function\\
${\bf x}_i\in{\mathbb R}^D, {\bf x}_j\in{\mathbb R}^D$& the $i$th and $j$th data point\\
${\cal S}_i, {\cal S}_j$& the $i$th and $j$th set\\
${\bf q}_i\in{\mathbb R}^D$& a query point\\
${\bf X}=\left [{\bf x}_1, \cdots, {\bf x}_N \right ]\in{\mathbb R}^{D\times N}$& data matrix with points as columns\\
${\bf y}_i\in\{1, -1\}^K$, or ${\bf y}_i\in\{0, 1\}^K$& hash codes of data points ${\bf x}_i$ and ${\bf x}_j$\\
${\bf y}_{k\cdot}\in{\mathbb B}^{N\times 1}$& the $k$-th hash bit of $N$ data points\\
${\bf Y}=\left [{\bf y}_1, \cdots, {\bf y}_N \right ]\in{\mathbb B}^{K\times N}$& hash codes of data $\bf X$\\
$\theta_{ij}$& angle between data points ${\bf x}_i$ and ${\bf x}_j$\\
${h}_k: {\bf R}^D\rightarrow \{1, -1\}$&the $k$-th hash function\\
$H=\left [h_1, \cdots, h_K \right ]: {\bf R}^D\rightarrow \{1, -1\}^{K}$& $K$ hash functions\\
$J({\cal S}_i, {\cal S}_j)$& Jaccard similarity between sets ${\cal S}_i$ and ${\cal S}_j$ \\
$J({\bf x}_i, {\bf x}_j)$& Jaccard similarity between vectors ${\bf x}_i$ and ${\bf x}_j$ \\
$d_{\cal H}({\bf y}_i, {\bf y}_j)$&Hamming distance between ${\bf y}_i$ and ${\bf y}_j$\\
$d_{\cal WH}({\bf y}_i, {\bf y}_j)$&weighted Hamming distance between ${\bf y}_i$ and ${\bf y}_j$\\
$({\bf x}_i, {\bf x}_j)\in{\cal M}$&a pair of similar points\\
$({\bf x}_i, {\bf x}_j)\in{\cal C}$&a pair of dissimilar points\\
$({\bf q}_i, {\bf x}_i^+, {\bf x}_i^-)$&a ranking triplet\\
$S_{ij}=sim({\bf x}_i, {\bf x}_j)$&similarity between data points ${\bf x}_i$ and ${\bf x}_j$\\
${\bf S}\in{\mathbb R}^{N\times N}$&similarity matrix of data ${\bf X}$\\
$\mathcal{P}_{\mathbf{w}}$&a hyperplane with its normal vector ${\bf w}$\\
\bottomrule
\end{tabular}
\label{Tab:notations}
\end{table*}

\subsection{Notations}
Given a sample point ${\bf x}\in {\mathbb R}^D$, one can employ a set of hash functions $H=\{h_1,\cdots,h_K\}$ to compute a $K$-bit binary code ${\bf y}=\{y_1,\cdots,y_K\}$ for ${\bf x}$ as
\begin{eqnarray}
\label{generalhashing}
{\bf y}=\{h_1({\bf x}), \cdots, h_2({\bf x}), \cdots, h_K({\bf x})\},
\end{eqnarray}
where the $k^{th}$ bit is computed as $y_k=h_k({\bf x})$. The hash function performs the mapping as $h_k: {\mathbb R}^D\longrightarrow{\mathbb B}$. Such a binary encoding process can also be viewed as mapping the original data point to a binary valued space, namely {\em Hamming space}:
\begin{eqnarray}
\label{Hammingmapping}
H: {\bf x}\rightarrow \{h_1({\bf x}), \cdots, h_K({\bf x})\}.
\end{eqnarray}
Given a set of hash functions, we can map all the items in the database ${\bf X}=\{{\bf x}_n\}_{n=1}^N\in{\mathbb R}^{D\times N}$ to the corresponding binary codes as $${\bf Y}=H({\bf X})=\{h_1({\bf X}),h_2({\bf X}),\cdots,h_K({\bf X})\},$$ where the hash codes of the data ${\bf X}$ are ${\bf Y}\in{\mathbb B}^{K\times N}$.

After computing the binary codes, one can perform ANN search in Hamming space with significantly reduced computation. Hamming distance between two binary codes ${\bf y}_i$ and ${\bf y}_j$ is defined as
\begin{eqnarray}
\label{sumhammingdistance}
d_{\cal H}({\bf y}_i,{\bf y}_j)=|{\bf y}_i-{\bf y}_j|=\sum_{k=1}^{K}|h_k({\bf x}_i)-h_k({\bf x}_j)|,
\end{eqnarray}
where ${\bf y}_i=\left [h_1({\bf x}_i), \cdots, h_k({\bf x}_i), \cdots, h_K({\bf x}_i) \right ]$ and ${\bf y}_j~=~\left [h_1({\bf x}_j), \cdots, h_k({\bf x}_j), \cdots, h_K({\bf x}_j)\right ]$. Note that the Hamming distance can be calculated in an efficient way as a bitwise logic operation. Thus, even conducting exhaustive search in the Hamming space can be significantly faster than doing the same in the original space. Furthermore, through designing a certain indexing structure, the ANN search with hashing methods can be even more efficient. Below we describe the pipeline of a typical hashing-based ANN search system.

\subsection{Pipeline of Hashing-based ANN Search}
There are three basic steps in ANN search using hashing techniques: designing hash functions, generating hash codes and indexing the database items, and online querying using hash codes. These steps are described in detail below.

\subsubsection{Designing Hash Functions}
There exist a number of ways of designing hash functions. Randomized hashing approaches often use random projections or permutations. The emerging {\em learning to hash} framework exploits the data distribution and often various levels of supervised information to determine optimal parameters of the hash functions. The supervised  information includes pointwise labels, pairwise relationships, and ranking orders. Due to their efficiency, the most commonly used hash functions are of the form of a generalized linear projection:
\begin{eqnarray}
\label{Hashfunctions}
h_k(x)=\sgn\left (f({\bf w}_k^\top {\bf x}+b_k)\right ).
\end{eqnarray}
Here $f(\cdot)$ is a prespecified function which can be possibly nonlinear. The parameters to be determined are $\{{\bf w}_k, b_k\}_{k=1}^K$, representing the projection vector ${\bf w}_k$ and the corresponding intercept $b_k$. During the training procedure, the data ${\bf X}$, sometimes along with supervised information, is used to estimate these parameters. In addition, different choices of $f(\cdot)$ yield different properties of the hash functions, leading to a wide range of hashing approaches. For example, LSH keeps $f(\cdot)$ to be an identity function, while shift-invariant kernel-based hashing and spectral hashing choose $f(\cdot)$ to be a shifted cosine or sinusoidal function~\cite{Weiss:2008NIPS}\cite{Raginsky:2009NIPS}.

Note that, the hash functions given by (\ref{Hashfunctions}) generate the codes as $h_k({\bf x)}\in\{-1, 1\}$. One can easily convert them into binary codes from $\{0, 1\}$ as
\begin{eqnarray}
\label{coverthashing}
y_k=\frac{1}{2}\left (1+h_k({\bf x})\right ).
\end{eqnarray}

\begin{figure}[t]
\begin{center}
\centerline{\includegraphics[width=0.99\columnwidth]{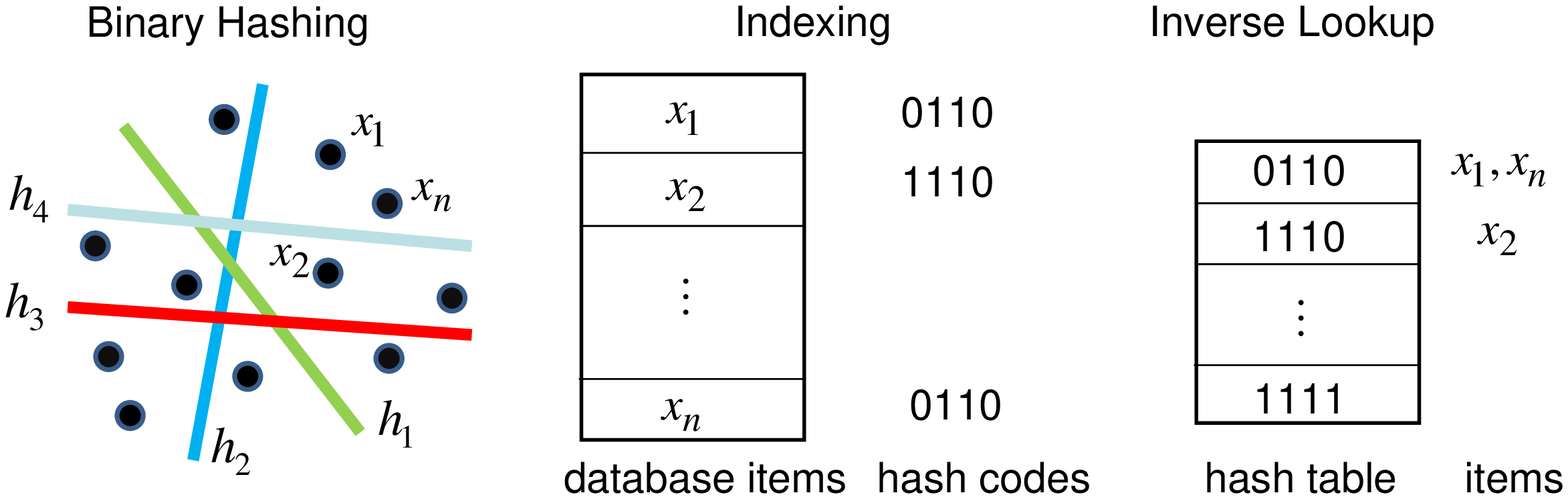}}
\caption{An illustration of linear projection based binary hashing, indexing, and hash table construction for fast inverse lookup.}
\label{fig:hashdigram}
\end{center}
\end{figure}

Without loss of generality, in this survey we will use the term {\it hash codes} to refer to either $\{0, 1\}$ or $\{-1, 1\}$ form, which should be clear from the context. 

\subsubsection{Indexing Using Hash Tables}
With a learned hash function, one can compute the binary codes ${\bf Y}$ for all the items in a database. For $K$ hash functions, the codes for the entire database cost only $NK/8$ bytes. Assuming the original data to be stored in double-precision floating-point format, the original storage costs $8ND$ bytes. Since the massive datasets are often associated with thousands of dimensions, the computed hash codes significantly reduce the storage cost by hundreds and even thousands of times.

In practice, the hash codes of the database are organized as an inverse-lookup, resulting in a {\em hash table} or a {\em hash map}. For a set of $K$ binary hash functions, one can have at most $2^K$ entries in the hash table. Each entry, called a hash bucket, is indexed by a $K$-bit hash code. In the hash table, one keeps only those buckets that contains at least one database item. Figure~\ref{fig:hashdigram} shows an example of using binary hash functions to index the data and construct a hash table. Thus, a hash table can be seen as an {\em inverse-lookup} table, which can return all the database items corresponding to a certain code in constant  time. This procedure is key to achieving speedup by many hashing based ANN search techniques. Since most of the buckets from $2^K$ possible choices are typically empty, creating an inverse lookup can be a very efficient way of even storing the codes if multiple database items end up with the same codes.

\subsubsection{Online Querying with Hashing}
\begin{figure}[t]
\begin{center}
\centerline{\includegraphics[width=0.72\columnwidth,height=0.36\columnwidth]{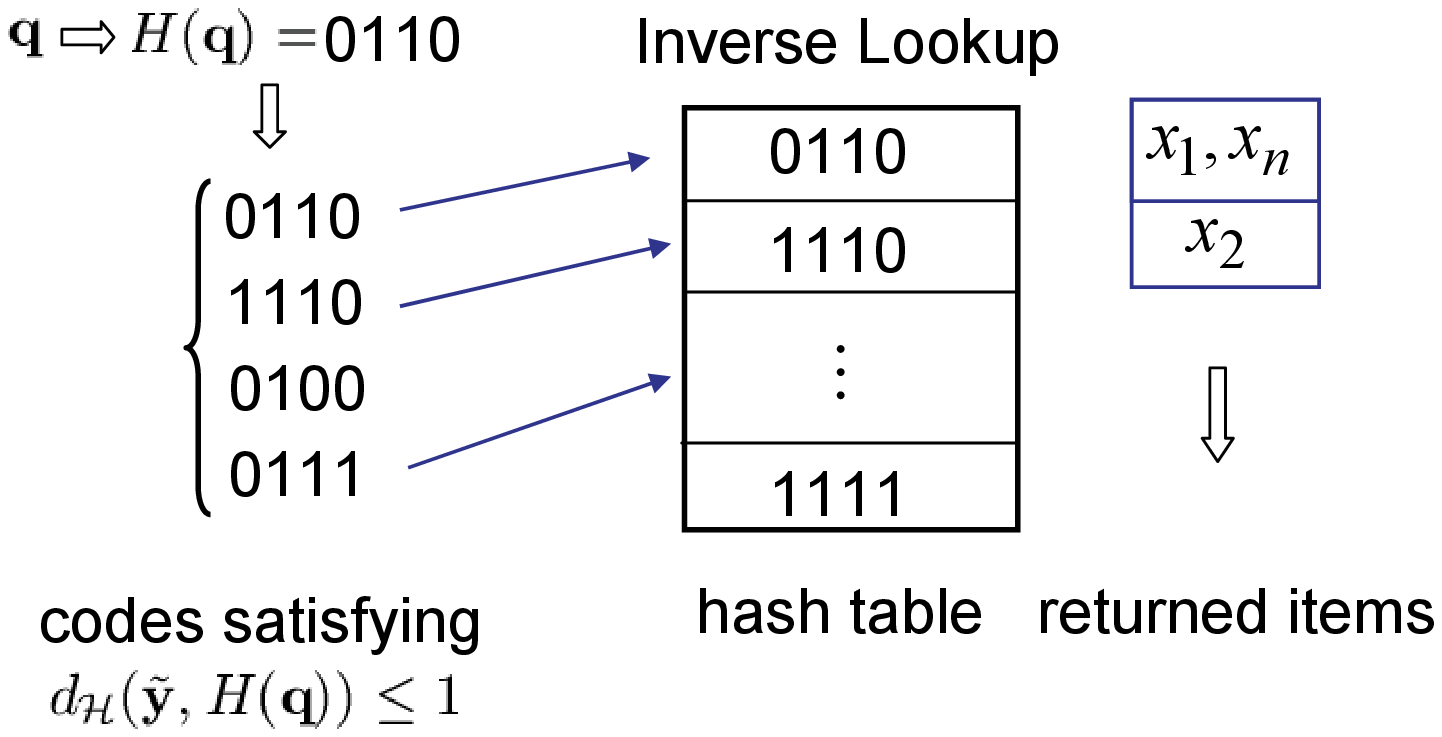}}
\caption{The procedure of inverse-lookup in hash table, where ${\bf q}$ is the query mapped to a 4-bit hash code ``0100'' and the returned approximate nearest neighbors within Hamming radius $1$ are ${\bf x}_1, {\bf x}_2, {\bf x}_n$.}
\label{fig:hashquery}
\end{center}
\end{figure}

During the querying procedure, the goal is to find the nearest database items to a given query. The query is first converted into a code using the same hash functions that mapped the database items to codes. One way to find nearest neighbors of the query is by computing the Hamming distance between the query code to all the database codes. Note that the Hamming distance can be rapidly computed using logical xor operation between binary codes as
\begin{eqnarray}
\label{xorhammingdistance}
d_{\cal H}({\bf y}_i,{\bf y}_j)={\bf y}_i\oplus{\bf y}_j.
\end{eqnarray}
On modern computer architectures, this is achieved efficiently by running xor instruction followed by popcount. 
With the computed Hamming distance between the query and each database item, one can perform exhaustive scan to extract the approximate nearest neighbors of the query. Although this is much faster than the exhaustive search in the original feature space, the time complexity is still linear. An alternative way of searching for the neighbors is by using the inverse-lookup in the hash table and returning the data points within a small Hamming distance $r$ of the query. Specifically, given a query point ${\bf q}$, and its corresponding hash code ${\bf y}_q=H({\bf q})$, all the database points ${\tilde {\bf y}}$ whose hash codes fall within the Hamming ball of radius $r$ centered at ${\bf y}_q$, i.,e. $d_{\cal H}({\tilde {\bf y}},H({\bf q}))\le r$. As shown in Figure~\ref{fig:hashquery}, for a $K$-bit binary code, a total of $\sum_{l=0}^{r}{{K}\choose{l}}$ possible codes will be within Hamming radius of $r$. Thus one needs to search $O(K^r)$ buckets in the hash table. The union of all the items falling into the corresponding hash buckets is returned as the search result. The inverse-lookup in a hash table has constant time complexity independent of the database size $N$. In practice, a small value of $r$ ($r=1, 2$ is commonly used) is used to avoid the exponential growth in the possible code combinations that need to be searched.

\subsection{Randomized Hashing Methods}

Randomized hashing, e.g. locality sensitive hash family, has been a popular choice due to its simplicity. In addition, it has interesting proximity preserving properties. A binary hash function $h(\cdot)$ from {\it LSH} family is chosen such that the probability of two points having the same bit is proportional to their (normalized) similarity, i.e.,
\begin{eqnarray}
\label{LSH}
P\left \{h({\bf x}_i)=h({\bf x}_j)\right \}=sim({\bf x}_i,{\bf x}_j).
\end{eqnarray}
Here $sim(\cdot, \cdot)$ represents similarity between a pair of points in the input space, e.g.,  cosine similarity or Jaccard similarity~\cite{Charikar:2002STOC}. In this section, we briefly review two categories of randomized hashing methods, i.e. random projection based and random permutation based approaches.

\subsubsection{Random Projection Based Hashing}
As a representative member of the LSH family, random projection based hash (RPH) functions have been widely used in different applications. The key ingredient of RPH is to map nearby points in the original space to the same hash bucket with a high probability. This equivalently preserves the locality in the original space in the Hamming space. Typical examples of RPH functions consist of a random projection ${\bf w}$ and a random shift $b$ as

\begin{eqnarray}
\label{RHHfunctions}
h_k(x)=\sgn({\bf w}_k^\top {\bf x}+b_k),
\end{eqnarray}
The random vector ${\bf w}$ is constructed by sampling each component of ${\bf w}$ randomly from a standard Gaussian distribution for cosine distance~\cite{Charikar:2002STOC}.

\begin{figure}[t]
	\begin{center}
		\centerline{\includegraphics[width=0.5\columnwidth,height=0.4\columnwidth]{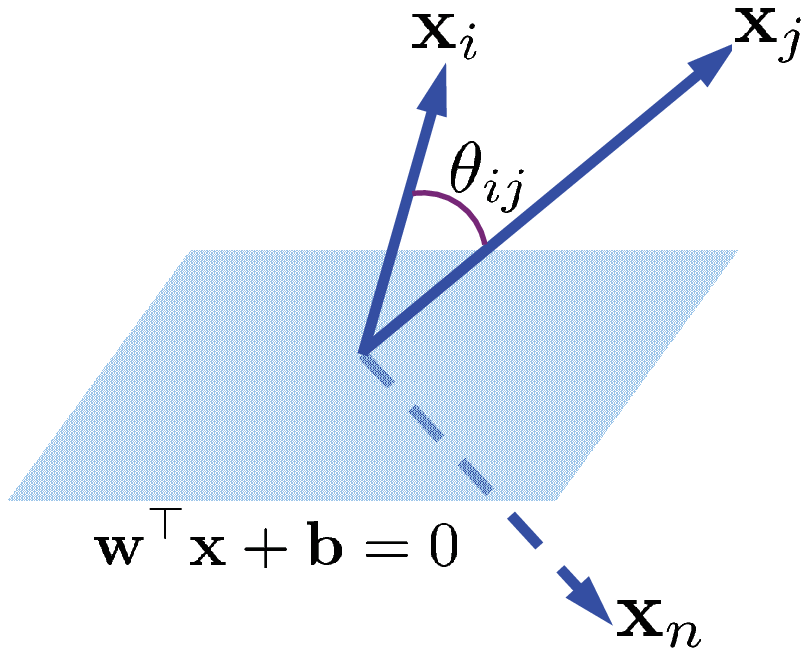}}
		\caption{An illustration of random hyperplane partitioning based hashing method.}
		\label{fig:randomhyperplanehashing}
	\end{center}
\end{figure}

It is easy to show that the collision probability of two samples ${\bf x}_i, {\bf x}_j$ falling into the same hash bucket is determined by the angle $\theta_{ij}$ between these two sample vectors, as shown in Figure~\ref{fig:randomhyperplanehashing}. One can show that
\begin{eqnarray}
\label{RHHcollision}
{\textrm Pr}\left [h_k({\bf x}_i)=h_k({\bf x}_j)\right ]=1-\frac{\theta_{ij}}{\pi}=1-\frac{1}{\pi}\cos^{-1}\frac{{\bf x}_i^\top{\bf x}_j}{\|{\bf x}_i\|\|{\bf x}_j\|}
\end{eqnarray}
The above collision probability gives the asymptotic theoretical guarantees for approximating the cosine similarity defined in the original space. However, long hash codes are required to achieve sufficient discrimination for high precision. This significantly reduces the recall if hash table based inverse lookup is used for search. In order to balance the tradeoff of precision and recall, one has to construct multiple hash tables with long hash codes, which increases both storage and computation costs. In particular, with hash codes of length $K$, it is required to construct a sufficient number of hash tables to ensure the desired performance bound~\cite{Datar:2004locality}. Given $l$ $K$-bit tables, the collision probability is given as:
\begin{eqnarray}
\label{LSHtablefunctions}
P\left \{ H({\bf x}_i)=H({\bf x}_j)\right \}\propto~~ l\cdot \left [1-\frac{1}{\pi}\cos^{-1}\frac{{\bf x}_i^\top{\bf x}_j}{\|{\bf x}_i\|\|{\bf x}_j\|}\right ]^K
\end{eqnarray}
To balance the search precision and recall, the length of hash codes should be long enough to reduce false collisions (i.e., non-neighbor samples falling into the same bucket). Meanwhile, the number of hash tables $l$ should be sufficiently large to increase the recall. However, this is inefficient due to extra storage cost and longer query time.

To overcome these drawbacks, many practical systems adapt various strategies to reduce the storage overload and to improve the efficiency. For instance, a self-tuning indexing technique, called {\it LSH} forest was proposed in~\cite{Bawa:2005WWW}, which aims at improving the performance without additional storage and query overhead. In~\cite{LV:2007VLDB}\cite{Dong:2008CIKM}, a technique called MultiProbe {\it LSH} was developed to reduce the number of required hash tables through intelligently probing multiple buckets in each hash table. In~\cite{FLSH:2011KDD}, nonlinear randomized Hadamard transforms were explored to speed up the LSH based ANN search for Euclidean distance. In~\cite{Satuluri:2012Bayesian}, BayesLSH was proposed to combine Bayesian inference with LSH in a principled manner, which has probabilistic guarantees on the quality of the search results in terms of accuracy and recall. However, the random projections based hash functions ignore the specific properties of a given data set and thus the generated binary codes are data-independent, which leads to less effective performance compared to the learning based methods to be discussed later.

In machine learning and data mining community, recent methods tend to leverage data-dependent and task-specific information to improve the efficiency of random projection based hash functions~\cite{Cayton:2007NIPS}. For example, incorporating kernel learning with {\it LSH} can help generalize ANN search from a standard metric space to a wide range of similarity functions~\cite{Kulis:2009ICCV}\cite{Mu:2010CVPR}. Furthermore, metric learning has been combined with randomized {\it LSH} functions to explore a set of pairwise similarity and dissimilarity constraints~\cite{Kulis:2009PAMI}. Other variants of locality sensitive hashing techniques include super-bit LSH~\cite{SBLSH:2012NIPS}, boosted LSH~\cite{Shakhnarovich:2005learning}, as well as non-metric LSH~\cite{NMLSH:2010AAAI}

\subsubsection{Random Permutation based Hashing}
Another well-known paradigm from the LSH family is min-wise independent permutation hashing (MinHash), which has been widely used for approximating Jaccard similarity between sets or vectors. Jaccard is a popular choice for measuring similarity between documents or images. A typical application is to index documents and then identify near-duplicate samples from a corpus of documents~\cite{Broder:1997CCS}\cite{MinPH:1998SODA}. The Jaccard similarity between two sets ${\cal S}_i$ and ${\cal S}_j$ is defined as $J({\cal S}_i, {\cal S}_j)=\frac{{\cal S}_i\cap{\cal S}_j}{{\cal S}_i\cup{\cal S}_j}$. Since a collection of sets $\{{\cal S}_i\}_{i=1}^N$ can be represented as a characteristic matrix ${\bf C}\in \mathbb{B}^{M\times N}$, where $M$ is the cardinality of the universal set ${\cal S}_1~\cup~\cdots~\cup{\cal S}_N$. Here the rows of ${\bf C}$ represents the elements of the universal set and the columns correspond to the sets. The element $c_{di}=1$ indicates the $d$-th element is a member of the $i$-th set, $c_{di}=0$ otherwise. Assume a random permutation $\pi_k(\cdot)$ that assigns the index of the $d$-th element as $\pi_k(d)\in\{1, \cdots, D\}$. It is easy to see that the random permutation satisfies two properties: $\pi_k(d)\neq \pi_k(l)$ and $Pr[\pi_k(d)>\pi_k(l)]=0.5$. A random permutation based min-hash signature of a set ${\cal S}_i$ is defined as the minimum index of the non-zero element after performing permutation using $\pi_k$
\begin{eqnarray}
\label{minhashsignature}
h_k({\cal S}_i)=\min_{d\in\{1, \cdots, D\}, c_{\pi_k(d)i}=1}\pi_k(d).
\end{eqnarray}
Note that such a hash function holds a property that the chance of two sets having the same MinHash values is equal to the Jaccard similarity between them~\cite{MMD:2012BOOK}
\begin{eqnarray}
\label{minhashproperty}
Pr[h_k({\cal S}_i)=h_k({\cal S}_j)]=J({\cal S}_i, {\cal S}_j).
\end{eqnarray}
The definition of the Jaccard similarity can be extended to two vectors ${\bf x}_i=\{x_{i1}, \cdots, x_{id}, \cdots, x_{iD}\}$ and ${\bf x}_j=\{x_{j1}, \cdots, x_{jd}, \cdots, x_{jD}\}$ as $$J({\bf x}_i, {\bf x}_j)=\frac{\sum_{d=1}^D\min(x_{id}, x_{jd})}{\sum_{d=1}^D\max(x_{id}, x_{jd})}.$$ Similar min-hash functions can be defined for the above vectors and the property of the collision probability shown in Eq.~\ref{minhashproperty} still holds \cite{Ioffe:2010ICDM}. Compared to the random projection based LSH family, the min-hash functions generate non-binary hash values that can be potentially extended to continuous cases. In practice, the min-hash scheme has shown powerful performance for high-dimensional and sparse vectors like the bag-of-word representation of documents or feature histograms of images. In a large scale evaluation conducted by Google Inc., the min-hash approach outperforms other competing methods for the application of webpage duplicate detection~\cite{Henzinger:2006SIGIR}. In addition, the min-hash scheme is also applied for Google news personalization~\cite{Das:2007WWW} and near duplicate image detection~\cite{NDID:2008BMVC}~\cite{PMH:2010ECCV}. Some recent efforts have been made to further improve the min-hash technique, including $b$-bit minwise hashing~\cite{BBMH:2010WWW}~\cite{BBMH:2010NIPS}, one permutation approach~\cite{OPH:2012NIPS}, geometric min-Hashing~\cite{GMH:2009CVPR}, and a fast computing technique for image data~\cite{FCMH:cvpr2012}.

\section{Categories of Learning Based Hashing Methods}
\label{sec:overview}
Among the three key steps in hashing-based ANN search, design of improved data-dependent hash functions has been the focus in {\em learning to hash} paradigm. Since the proposal of LSH in~\cite{Indyk:1998STOC}, many new hashing techniques have been developed. Note that most of the emerging hashing methods are focused on improving the search performance using a single hash table. The reason is that these techniques expect to learn compact discriminative codes such that searching within a small Hamming ball of the query or even exhaustive scan in Hamming space is both fast and accurate.   Hence, in the following, we primarily focus on various techniques and algorithms for designing a single hash table. In particular, we provide different perspectives such as the learning paradigm and hash function characteristics to categorize the hashing approaches developed recently. It is worth mentioning that a few recent studies have shown that exploring the power of multiple hash tables can sometimes generate superior performance. In order to improve precision as well as recall, Xu et al., developed multiple complementary hash tables that are sequentially learned using a boosting-style algorithm~\cite{Xu:2011ICCV}. Also, in cases when the code length is not very large and the number of database points is large, exhaustive scan in Hamming space can be done much faster by using multi-table indexing as shown by Norouzi et al.~\cite{MIH:2012CVPR}.

\subsection{Data-Dependent vs. Data-Independent}
Based on whether design of hash functions requires analysis of a given dataset, there are two high-level categories of hashing techniques: data-independent and data-dependent. As one of the most popular data-independent approaches, random projection has been used widely for designing data-independent hashing techniques such as LSH and SIKH mentioned earlier. LSH is arguably the most popular hashing method and has been applied to a variety of  problem domains, including information retrieval and computer vision. In both LSH and SIKH, the projection vector ${\bf w}$ and intersect $b$, as defined in Eq.~\ref{Hashfunctions}, are randomly sampled from certain distributions. Although these methods have strict performance guarantees, they are less efficient since the hash functions are not specifically designed for a certain dataset or search task. Based on the random projection scheme, there have been several efforts to improve the performance of the LSH method~\cite{LV:2007VLDB}~\cite{FLSH:2011KDD}~\cite{Kulis:2009ICCV}.

Realizing the limitation of data-independent hashing approaches, many recent methods use data and possibly some form of  supervision to design more efficient hash functions. Based on the level of supervision, the data-dependent methods can be further categorized into three subgroups as described below.

\subsection{Unsupervised, Supervised, and Semi-Supervised}
Many emerging hashing techniques are designed by exploiting various machine learning paradigms, ranging from unsupervised and supervised to semi-supervised settings. For instance, unsupervised hashing methods attempt to integrate the data properties, such as distributions and manifold structures to design compact hash codes with improved accuracy. Representative unsupervised methods include spectral hashing~\cite{Weiss:2008NIPS}, graph hashing~\cite{GH:icml2011}, manifold hashing~\cite{IHM:cvpr2013}, iterative quantization hashing~\cite{IQ:2011CVPR}, kernalized locality sensitive hashing~\cite{Kulis:2009ICCV}\cite{Kulis:2011PAMI}, isotropic hashing~\cite{IH:2012NIPS}, angular quantization hashing~\cite{AQ:2012NIPS}, and spherical hashing~\cite{SH:2012CVPR}. Among these approaches, spectral hashing explores the data distribution and graph hashing utilizes the underlying manifold  structure of data captured by a graph representation. In addition, supervised learning paradigms ranging from kernel learning to metric learning to deep learning have been exploited to learn binary codes, and many supervised hashing methods have been proposed recently~\cite{Kulis:2009PAMI}\cite{SKH:cvpr2012}\cite{Norouzi:NIPS2012}\cite{Torralba:2008CVPR}\cite{SBHJSD:2013ICCV}\cite{SDH:CVPR2015}. Finally, semi-supervised learning paradigm was employed to design hash functions by using both labeled and unlabeled data. For instance, Wang et. al proposed a regularized objective to achieve accurate yet balanced hash codes to avoid overfitting~\cite{SSH:2012PAMI}. In~\cite{Jain:2008CVPR}\cite{Jain:2008NIPS}, authors proposed to exploit the metric learning and locality sensitive hashing to achieve fast similarity based search. Since the labeled data is used for deriving optimal metric distance while the hash function design uses no supervision, the proposed hashing technique can be regarded as a semi-supervised approach.

\begin{figure}[t]
 \centering
 \subfigure[ ]
 {\label{fig:differentsupervision:a}
 \includegraphics[width=0.31\linewidth]{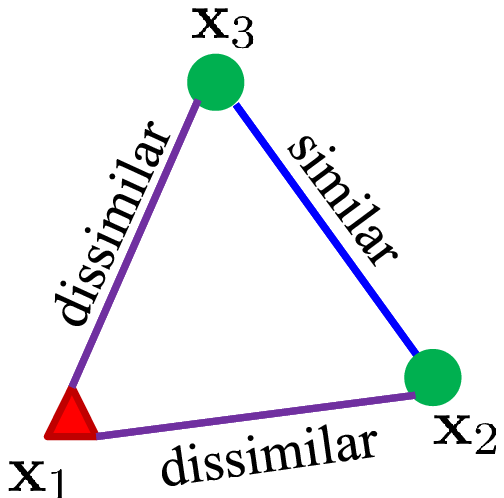}}
  \subfigure[ ]
 {\label{fig:differentsupervision:b}
 \includegraphics[width=0.31\linewidth]{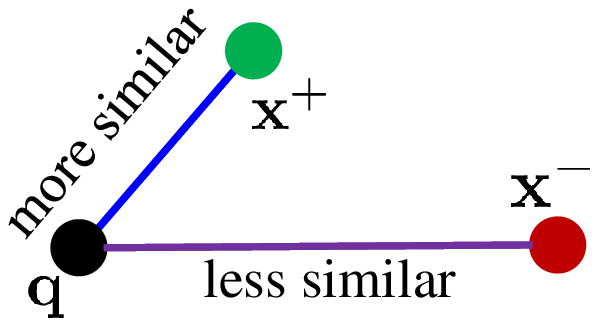}}
  \subfigure[ ]
 {\label{fig:differentsupervision:c}
 \includegraphics[width=0.31\linewidth]{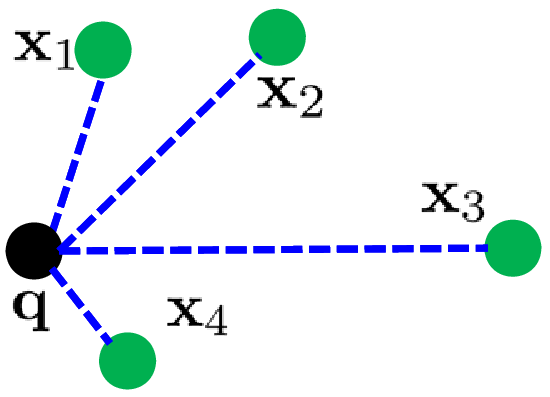}}
 \caption{An illustration of different levels of supervised information: a) pairwise labels; b) a triplet $sim({\bf q}, {\bf x}^+)>sim({\bf q}, {\bf x}^-)$; and c) a distance based rank list $({\bf x}_4, {\bf x}_1, {\bf x}_2, {\bf x}_3)$ to a query point $\bf q$.}
\label{fig:differentsupervision}
\end{figure}

\subsection{Pointwise, Pairwise, Triplet-wise and Listwise}
Based on the level of supervision, the supervised or semi-supervised hashing methods can be further grouped into several subcategories, including {\em pointwise}, {\em pairwise}, {\em triplet-wise}, and {\em listwise} approaches. For example, a few existing approaches utilize the instance level semantic attributes or labels to design the hash functions~\cite{Shakhnarovich:2003ICCV}\cite{Shakhnarovich:2005learning}\cite{PDBC:2012ECCV}. Additionally, learning methods based on pairwise supervision have been extensively studied, and many hashing techniques have been proposed~\cite{Kulis:2009PAMI}\cite{SKH:cvpr2012}\cite{Norouzi:NIPS2012}\cite{Torralba:2008CVPR}\cite{SSH:2012PAMI}\cite{Jain:2008CVPR}\cite{Jain:2008NIPS}\cite{NTH:2015KDD}. As demonstrated in Figure~\ref{fig:differentsupervision:a}, the pair $({\bf x}_2, {\bf x}_3)$ contains similar points and the other two pairs $({\bf x}_1, {\bf x}_2)$ and $({\bf x}_1, {\bf x}_3)$ contain dissimilar points. Such relations are considered in the learning procedure to preserve the pairwise label information in the learned Hamming space. Since the ranking information is not fully utilized, the performance of pairwise supervision based methods could be sub-optimal for nearest neighbor search. More recently, a triplet ranking that encodes the pairwise proximity comparison among three data points is exploited to design hash codes~\cite{CG:2013ICML}\cite{Norouzi:NIPS2012}\cite{Wang:2013ACMMM}. As shown in Figure~\ref{fig:differentsupervision:b}, the point ${\bf x}^+$ is more similar to the query point ${\bf q}$ than the point ${\bf x}^-$. Such a triplet ranking information, i.e., $sim({\bf q}, {\bf x}^+)>sim({\bf q}, {\bf x}^-)$ is expected to be encoded in the learned binary hash codes. Finally, the listwise information indicates the rank order of a set of points with respect to the query point. In Figure~\ref{fig:differentsupervision:c}, for the query point ${\bf q}$, the rank list $({\bf x}_4, {\bf x}_1, {\bf x}_2, {\bf x}_3)$ shows the ranking order of their similarities to the query point $\bf q$, where ${\bf x}_4$ is the nearest point and ${\bf x}_3$ is the farthest one. By converting rank lists to a triplet tensor matrix, listwise hashing is designed to preserve the ranking in the Hamming space~\cite{ListwiseH:2013ICCV}.

\subsection{Linear vs. Nonlinear}
\begin{figure}[t]
	\centering
	\subfigure[ ]
	{\label{fig:PCA_SH_Projection:a}
		\includegraphics[width=0.48\linewidth]{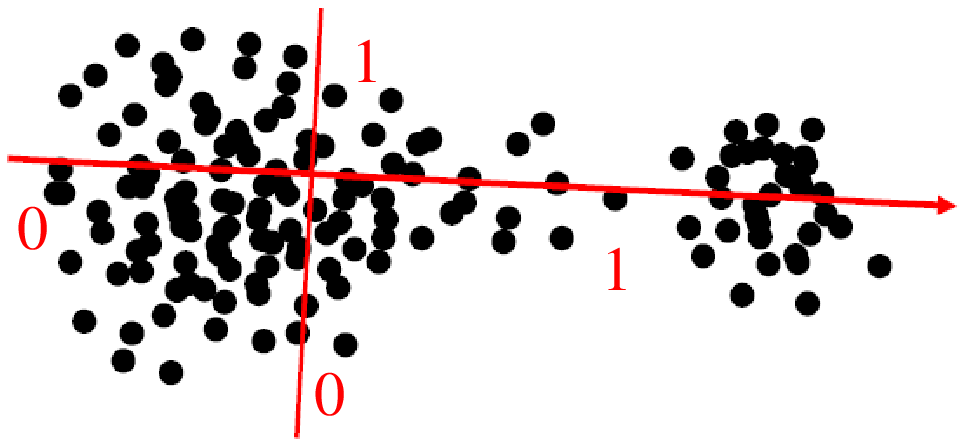}}
	\subfigure[ ]
	{\label{fig:PCA_SH_Projection:b}
		\includegraphics[width=0.48\linewidth]{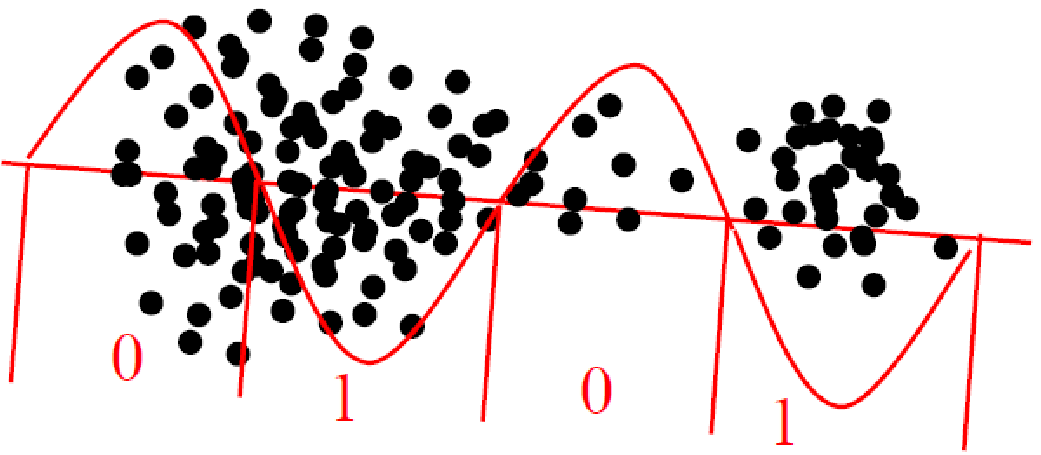}}
	\caption{Comparison of hash bits generated using a) PCA hashing and b) Spectral Hashing.}
	\label{fig:PCA_SH_Projection}
\end{figure}

Based on the form of function $f(\cdot)$ in Eq.~\ref{Hashfunctions}, hash functions can also be categorized in two groups: linear and nonlinear. Due to their computational efficiency, linear functions tend to be more popular, which include random projection based LSH methods. The learning based methods derive optimal projections by optimizing  different types of objectives. For instance, PCA hashing performs principal component analysis on the data to derive large variance projections~\cite{AQ:2012NIPS}\cite{ALD2010aggregating}\cite{ODH:2012ICML}, as shown in Figure~\ref{fig:PCA_SH_Projection:a}. In the same league, supervised methods have used Linear Discriminant Analysis to design more discriminative hash codes~\cite{LDAH:2012PAMI}\cite{EDPCB:2012ECCV}. Semi-supervised hashing methods estimate the projections that have minimum empirical loss on pair-wise labels while partitioning the unlabeled data in a balanced way~\cite{SSH:2012PAMI}. Techniques that use variance of the projections as the underlying objective, also tend to use orthogonality constraints for computational ease. However, these constraints lead to a significant drawback since the variance for most real-world data decays rapidly with most of the variance contained only in top few directions. Thus, in order to generate more bits in the code, one is forced to use progressively low-variance directions due to orthogonality constraints. The binary codes derived from these low-variance projections tend to have   significantly lower performance. Two types of solutions based on relaxation of the orthogonality constraints or random/learned rotation of the data have been proposed in the literature to address these issues~\cite{SSH:2012PAMI}\cite{IQ:2011CVPR}. Isotropic hashing is proposed to derive projections with equal variances and is shown to be superior to anisotropic variances based projections~\cite{IH:2012NIPS}. Instead of performing one-shot learning, sequential projection learning derives correlated projections with the goal of correcting errors from previous hash bits~\cite{SPLH:icml2010}. Finally, to reduce the computational complexity of full projection, circulant binary embedding was recently proposed to significantly speed up the encoding process using the circulant convolution~\cite{YU:icml2014}.

Despite its simplicity, linear hashing often suffers from insufficient discriminative power. Thus, nonlinear methods have been developed to override such limitations. For instance, spectral hashing first extracts the principal projections of the data, and then partitions the projected data by a sinusoidal function (nonlinear) with a specific angular frequency. Essentially, it prefers to partition projections with large spread and small spatial frequency such that the large variance projections can be reused. As illustrated in Figure~\ref{fig:PCA_SH_Projection:b}, the fist principal component can be reused in spectral hashing to divide the data into four parts while being encoded with only one bit. In addition, shift-invariant kernel-based hashing chooses $f(\cdot )$ to be a shifted cosine function and samples the projection vector in the same way as standard LSH does~\cite{Raginsky:2009NIPS}.  Another category of nonlinear hashing techniques employs kernel functions~\cite{Kulis:2011PAMI}\cite{SKH:cvpr2012}\cite{GH:icml2011}\cite{OKH:2010KDD}. Anchor graph hashing proposed by Liu et al.~\cite{GH:icml2011} uses a kernel function to measure similarity of each points with a set of anchors resulting in nonlinear hashing. Kernerlized LSH uses a sparse set of datapoints to compute a kernel matrix and preform random projection in the kernel space to compute binary codes~\cite{Kulis:2011PAMI}. Based on similar representation of kernel metric, Kulis and Darrell propose learning of hash functions by explicitly minimizing the reconstruction error in the kernel space and Hamming space~\cite{Kulis:2009NIPS}. Liu et al. applies kernel representation but optimizes the hash functions by exploring the equivalence between optimizing the code inner products and the Hamming distances to achieve scale invariance~\cite{SKH:cvpr2012}.

\begin{table*}[htbp]
\renewcommand{\arraystretch}{1.35}
\setlength{\tabcolsep}{3pt}
\centering
\begin{tabular}{|c|c|c|c|c|}
\hline
\small Method &\small Hash Function/Objective Function &\small Parameters&\small Learning Paradigm&\small Supervision\\
\hline\hline
\small {\it Spectral Hashing}&$\sgn(\cos(\alpha{\bf w}^\top {\bf x}))$&${\bf w}, \alpha$&\small unsupervised&NA\\\hline
\small {\it Anchor Graph Hashing}&$\sgn({\bf w}^\top {\bf x})$&{\bf w}&\small unsupervised&NA\\\hline
\small {\it Angular Quantization}&$({\bf b}, {\bf w})=\arg\max\sum_i\frac{{\bf b}_i^\top}{\|{\bf b}_i\|_2}{\bf w}^\top{\bf x}_i$&{\bf b}, {\bf w}&\small unsupervised&NA\\[0.1cm]\hline
\small {\it Binary Reconstructive Embedding}&$\sgn({\bf w}^\top K({\bf x}))$&${\bf w}$&\small unsupervised/supervised&\small pairwise distance\\\hline
\small {\it Metric Learning Hashing}&$\sgn({\bf w}^\top {\bf G}^\top{\bf x})$&${\bf G}, {\bf w}$&\small supervised&\small pairwise similarity\\ \hline
\small {\it Semi-Supervised Hashing}& $\sgn({\bf w}^\top {\bf x})$&${\bf w}$ &\small semi-supervised&\small pairwise similarity \\ \hline
\small {\it Column generation hashing}& $\sgn({\bf w}^\top {\bf x}+b)$&\small $\bf w$&\small supervised&\small triplet \\
\hline
\small {\it Listwise hashing}&$\sgn({\bf w}^\top {\bf x}+b)$&$\bf w$&\small supervised&\small ranking list\\
\hline
\small {\it Circulant binary embedding}&$\sgn(circ({\bf r})\cdot{\bf x})$&$\bf r$&\small unsupervised/supervised&\small pairwise similarity\\
\hline
\end{tabular}
\caption{A summary of the surveyed hashing techniques in this article.}
\label{Tab:conceptualcomparison}
\end{table*}

\subsection{Single-Shot Learning vs. Multiple-Shot Learning}
For learning based hashing methods, one first formulates an objective function reflecting desired characteristics of the hash codes. In a single-shot learning paradigm, the optimal solution is derived by optimizing the objective function in a single-shot. In such a learning to hash framework, the $K$ hash functions are learned simultaneously. In contrast, the multiple-shot learning procedure considers a global objective, but optimizes a hash function considering the bias generated by the previous hash functions. Such a procedure sequentially trains hash functions one bit at a time~\cite{SPEC:cvpr2010}\cite{SPLH:icml2010}\cite{CPH:2013ICCV}. The multiple-shot hash function learning is often used in supervised or semi-supervised settings since the given label information can be used to assess the quality of the hash functions learned in previous steps. For instance, the sequential projection based hashing aims to incorporate the bit correlations by iteratively updating the pairwise label matrix, where higher weights are imposed on point pairs violated by the previous hash functions~\cite{SPLH:icml2010}. In the complementary projection learning approach~\cite{CPH:2013ICCV}, the authors present a sequential learning procedure to obtain a series of hash functions that cross the sparse data region, as well as generate balanced hash buckets. Column generation hashing learns the best hash function during each iteration and updates the weights of hash functions accordingly. Other interesting learning ideas include two-step learning methods which treat hash bit learning and hash function learning separately~\cite{TSH:2013ICCV}\cite{STH:2010SIGIR}.


\subsection{Non-Weighted vs. Weighted Hashing}
Given the Hamming embedding defined in Eq.~\ref{Hammingmapping}, traditional hashing based indexing schemes map the original data into a {\em non-weighted} Hamming space, where each bit contributes equally. Given such a mapping, the Hamming distance is calculated by counting the number of different bits. However, it is easy to observe that different bits often behave differently~\cite{SSH:2012PAMI}\cite{Weiss:2008NIPS}. In general, for linear projection based hashing methods, the binary code generated from large variance projection tends to perform better due to its superior discriminative power. Hence, to improve discrimination among hash codes, techniques were designed to learn a weighted hamming embedding as
\begin{eqnarray}
\label{WeightedHammingMapping}
H: {\cal X}\rightarrow \{\alpha_1h_1({\bf x}), \cdots, \alpha_Kh_K({\bf x})\}.
\end{eqnarray}
Hence the conventional hamming distance is replaced by a weighted version as
\begin{eqnarray}
\label{weightedhammingdis}
d_{\cal WH}=\sum_{k=1}^{K}\alpha_k|h_k({\bf x}_i)-h_k({\bf x}_j)|.
\end{eqnarray}
One of the representative approaches is Boosted Similarity Sensitive Coding (BSSC)~\cite{Shakhnarovich:2005learning}. By learning the hash functions and the corresponding weights $\{\alpha_1, \cdots, \alpha_k\}$ jointly, the objective is to lower the collision probability of non-neighbor pair $({\bf x}_i, {\bf x}_j)\in{\cal C}$ while improving the collision probability of neighboring pair $({\bf x}_i, {\bf x}_j)\in {\cal M}$. If one treats each hash function as a decision stump, the straightforward way of learning the weights is to directly apply adaptive boosting algorithm~\cite{Freund:1995CLT} as described in~\cite{Shakhnarovich:2005learning}. In~\cite{BOOMAP:2008PAMI}, a boosting-style method called BoostMAP is proposed to map data points to weighted binary vectors that can leverage both metric and semantic similarity measures. Other weighted hashing methods include designing specific bit-level weighting schemes to improve the search accuracy~\cite{CG:2013ICML}\cite{Jiang:2013TMM}\cite{Jiang:2011ICMR}\cite{Wang:2013CIKM}\cite{BCR:cvpr2013}. In addition, a recent work about designing a unified bit selection framework can be regarded as a special case of weighted hashing approach, where the weights of hash bits are binary~\cite{HBS:cvpr2013}. Another effective hash code ranking method is the query-sensitive hashing, which explores the raw feature of the query sample and learns query-specific weights of hash bits to achieve accurate $\epsilon$-nearest neighbor search~\cite{QSRANK:2012CVPR}. 



%
%

%



\section{Methodology Review and Analysis}
\label{sec:methods}
In this section, we will focus on review of several representative hashing methods that explore various machine learning techniques to design data-specific indexing schemes. The techniques consist of unsupervised, semi-supervised, as well as supervised approaches, including spectral hashing, anchor graph hashing, angular quantization, binary reconstructive embedding based hashing, metric learning based hashing, semi-supervised hashing, column generation hashing, and ranking supervised hashing. Table~\ref{Tab:conceptualcomparison} summarizes the surveyed hashing techniques, as well as their technical merits. 

Note that this section mainly focuses on describing the intuition and formulation of each method, as well as discussing their pros and cons. The performance of each individual method highly depends on practical settings, including learning parameters and dataset itself. In general, the nonlinear and supervised techniques tend to generate better performance than linear and unsupervised methods, while being more computationally costly~\cite{SSH:2012PAMI}\cite{Kulis:2009PAMI}\cite{Kulis:2011PAMI}\cite{SKH:cvpr2012}\cite{Kulis:2009NIPS}.

\subsection{Spectral Hashing}
In the formulation of spectral hashing, the desired properties include keeping neighbors in input space as neighbors in the hamming space and requiring the codes to be balanced and uncorrelated~\cite{Weiss:2008NIPS}. Hence, the objective of spectral hashing is formulated as:
\begin{eqnarray}
\label{SHproperties}
\min \sum_{ij} \frac{1}{2}A_{ij} \|{\bf y}_i-{\bf y}_j\|^2=\frac{1}{2}\tr({\bf Y}^\top{\bf L}{\bf Y})\\ \nonumber
\textrm{subject to:}~~~~~~~~~~~~~~~~~~~~~~~~~~~ {\bf Y}\in \{-1, 1\}^{N\times K} \\ \nonumber
{\bf 1}^\top {\bf y}_{k\cdot}=0,~~k=1, \cdots, K \\ \nonumber\
{\bf Y}^\top{\bf Y}=n{\bf I}_{K\times K},
\end{eqnarray}
where ${\bf A}=\{A_{ij}\}_{i,j=1}^N$ is a pairwise similarity matrix and the Laplacian matrix is calculated as $\bf L=\diag({\bf A}{\bf 1})-{\bf A}$. The constraint ${\bf 1}^\top {\bf y}_k=0$ ensures that the hash bit ${\bf y}_k$ reaches a balanced partitioning of the data and the constraint ${\bf Y}^\top{\bf Y}=n{\bf I}_{K\times K}$ imposes orthogonality between hash bits to minimize the redundancy.

The direct solution for the above optimization is non-trivial for even a single bit since it is essentially a balanced graph partition problem, which is NP hard. The orthogonality constraints for $K$-bit balanced partitioning make the above problem even harder. Motivated by the well-known spectral graph analysis~\cite{Fowlkes:2004PAMI}, the authors suggest to minimize the cost function with relaxed constraints. In particular, with the assumption of uniform data distribution, the spectral solution can be efficiently computed using $1$D-Laplacian eigenfunctions~\cite{Weiss:2008NIPS}. The final solution for spectral hashing equals to apply a sinusoidal function with pre-computed angular frequency to partition data along PCA directions. Note that the projections are computed using data but learned in an unsupervised manner. As most of the orthogonal projection based hashing methods, spectral hashing suffers from the low-quality binary coding using low-variance projections. Hence, a ``kernel trick'' is used to alleviate the degraded performance when using long hash bits~\cite{MSH:2012ECCV}. Moreover, the assumption of uniform data distribution usually hardly hold for real-world data.

\subsection{Anchor Graph Hashing}
\begin{figure}[t]
 \centering
 \subfigure[ ]
 {\label{fig:twobitscomparision:a}
 \includegraphics[width=0.31\linewidth]{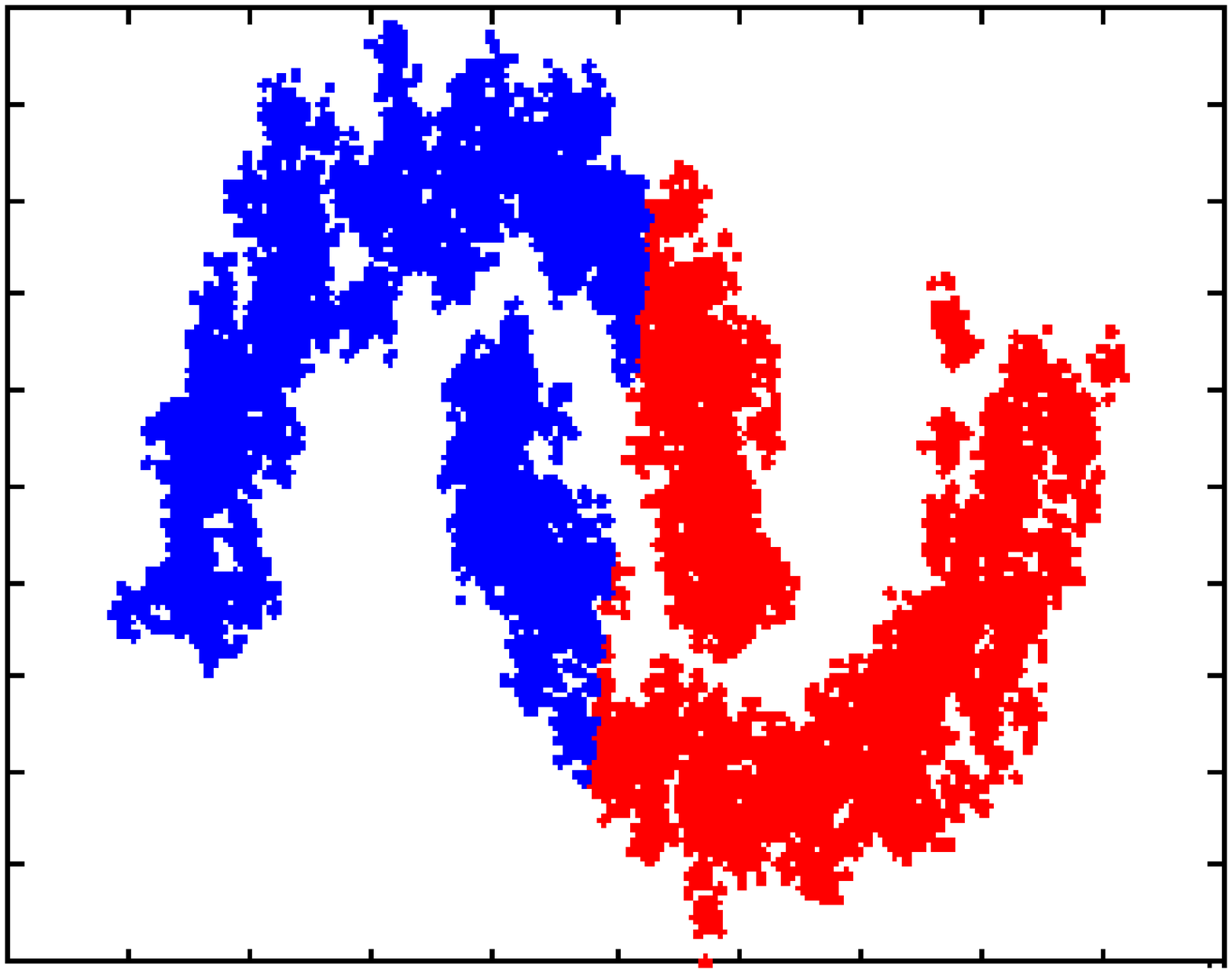}}
  \subfigure[ ]
 {\label{fig:twobitscomparision:b}
 \includegraphics[width=0.31\linewidth]{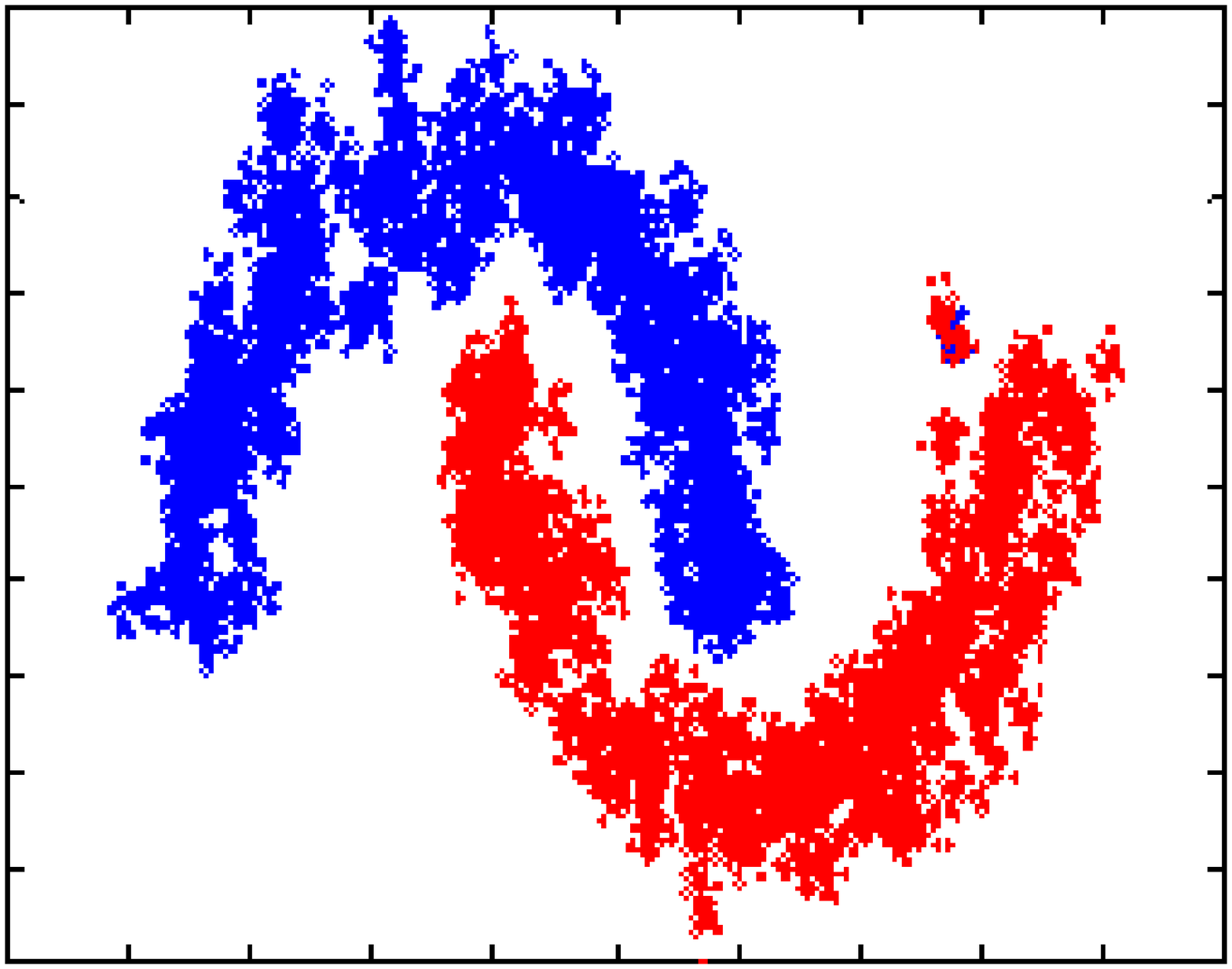}}
  \subfigure[ ]
 {\label{fig:twobitscomparision:c}
 \includegraphics[width=0.31\linewidth]{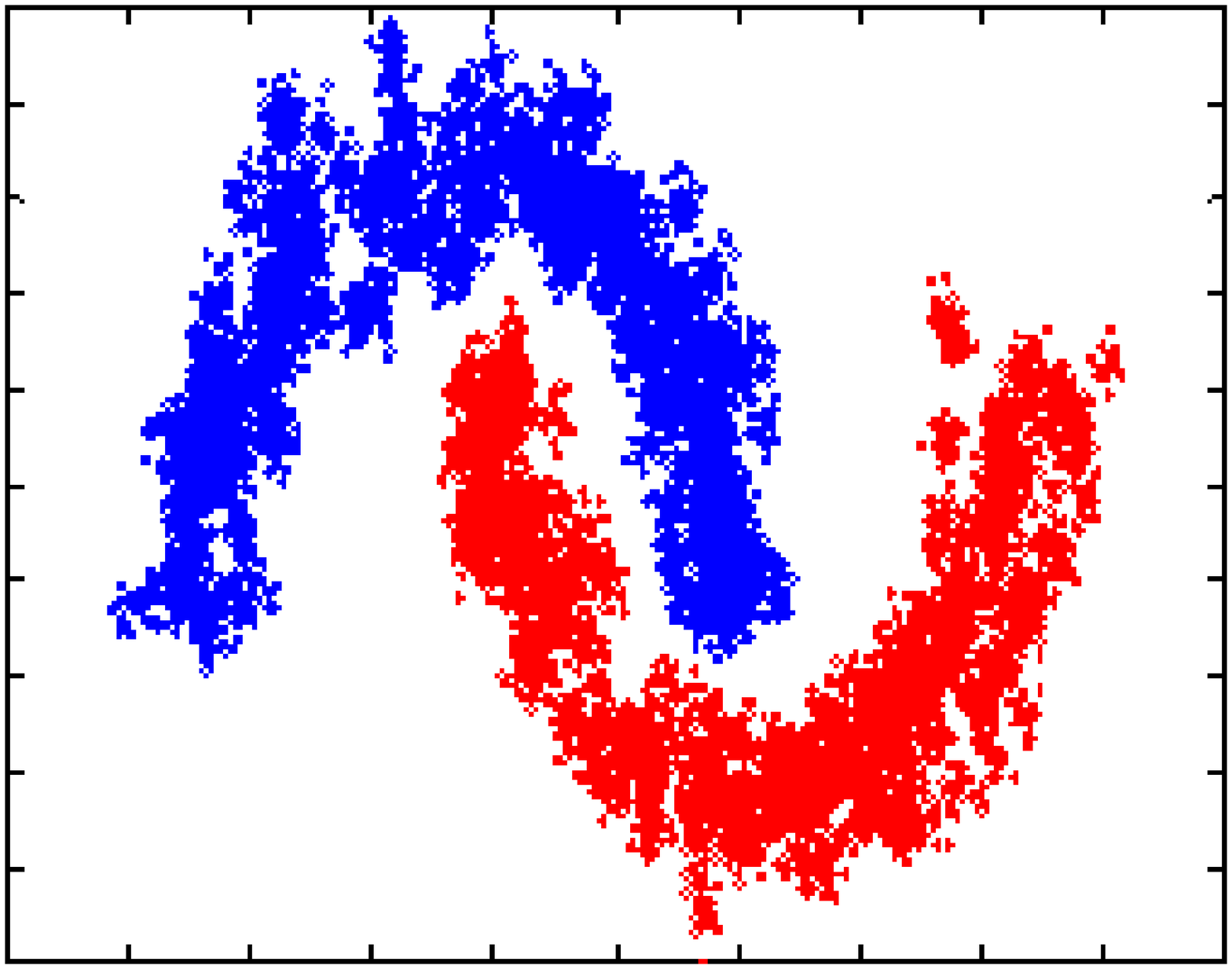}}
 \subfigure[ ]
 {\label{fig:twobitscomparision:d}
 \includegraphics[width=0.31\linewidth]{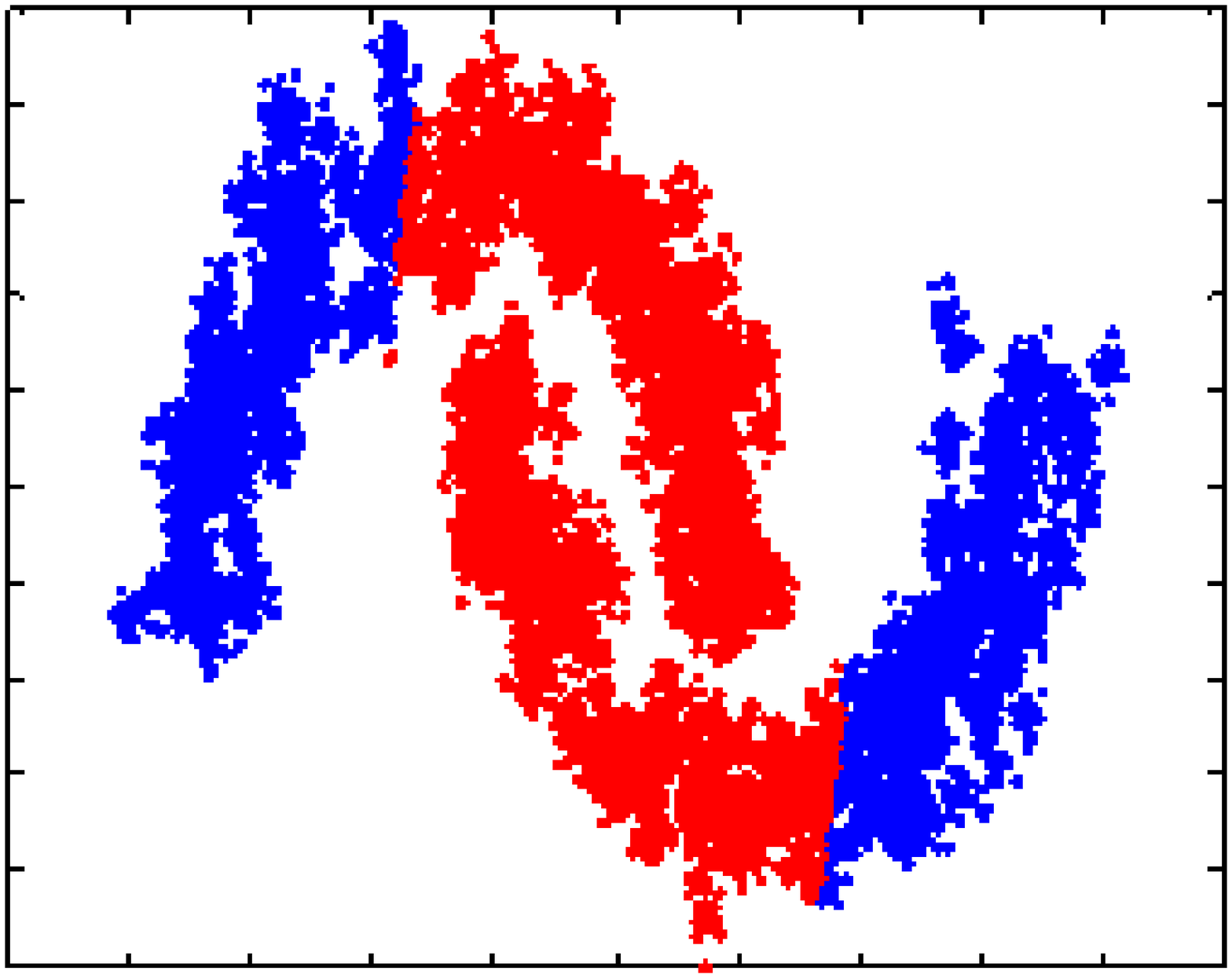}}
  \subfigure[ ]
 {\label{fig:twobitscomparision:e}
 \includegraphics[width=0.31\linewidth]{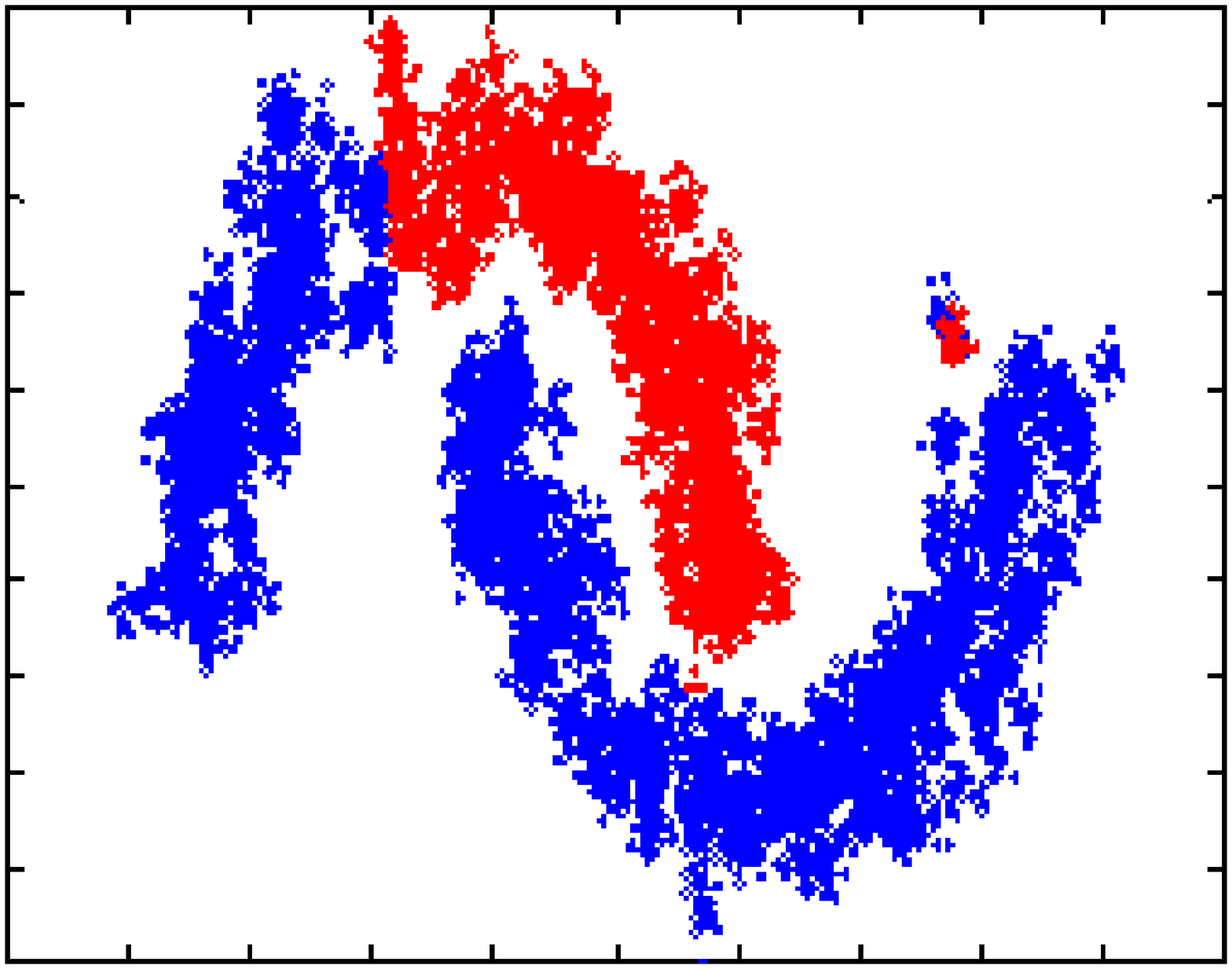}}
  \subfigure[ ]
 {\label{fig:twobitscomparision:f}
 \includegraphics[width=0.31\linewidth]{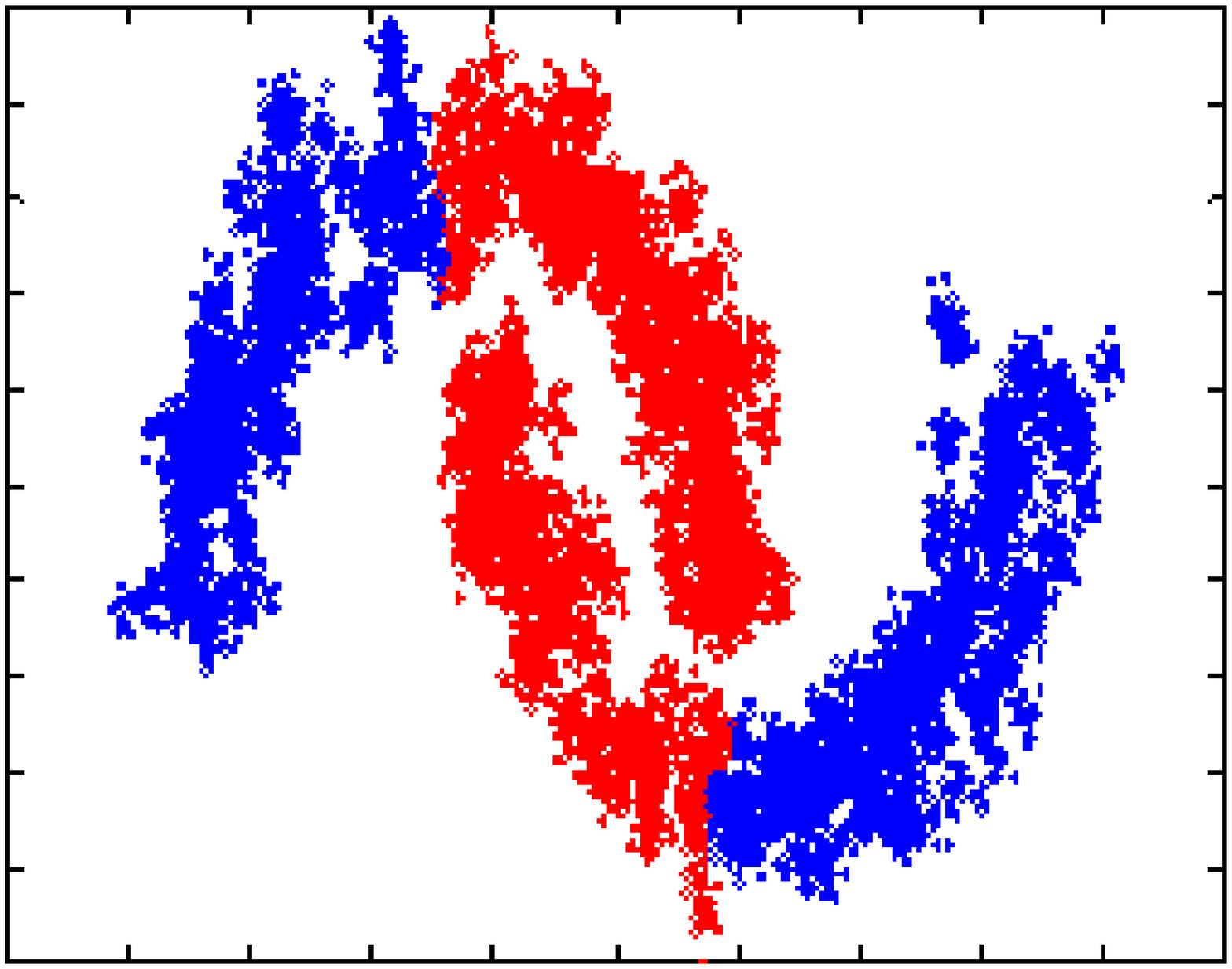}}
 \caption{Comparison of partitioning a two-moon data by the first two hash bits using different methods: a) the first bit using spectral hashing; b) the first bit using exact graph hashing; c) the first bit using anchor graph hashing; d) the second bit using spectral hashing; e) the second bit using exact graph hashing; f) the second bit using anchor graph hashing;}
\label{fig:twobitscomparision}
\end{figure}

Following the similar objective as spectral hashing, anchor graph hashing was designed to solve the problem from a different perspective without the assumption of uniform distribution~\cite{GH:icml2011}. Note that the critical bottleneck for solving Eq.~\ref{SHproperties} is the cost of building a pariwsie similarity graph ${\bf A}$, the computation of associated graph Laplacian, as well as solving the corresponding eigen-system, which at least has a quadratic complexity. The key idea is to use a small set of $M (M\ll N)$ anchor points to approximate the graph structure represented by the matrix ${\bf A}$ such that the similarity between any pair of points can be approximated using point-to-anchor similarities~\cite{AGC:2010ICML}. In particular, the truncated point-to-anchor similarity ${\bf Z}\in{\mathbb R}^{N\times M}$ gives the similarities between $N$ database points to the $M$ anchor points. Thus, the approximated similarity matrix ${\hat {\bf A}}$ can be calculated as ${\hat {\bf A}}={\bf Z}{\bf \Lambda}{\bf Z}^\top$, where ${\bf \Lambda}=\diag({\bf Z}{\bf 1})$ is the degree matrix of the anchor graph ${\bf Z}$. Based on such an approximation, instead of solving the eigen-system of the matrix ${\hat {\bf A}}={\bf Z}{\bf \Lambda}{\bf Z}^\top$, one can alternatively solve a much smaller eigen-system with an $M\times M$ matrix ${\bf \Lambda}^{1/2}{\bf Z}^\top{\bf Z}{\bf \Lambda}^{1/2}$. The final binary codes can be obtained through calculating the sign function over a spectral embedding as
\begin{eqnarray}
\label{AGHproperties}
{\bf Y}=\sgn({\bf Z}{\bf \Lambda}^{1/2}{\bf V}{\bf \Sigma}^{1/2}),
\end{eqnarray}
Here we have the matrices ${\bf V}=[{\bf v}_1, \cdots, {\bf v}_k, \cdots, {\bf v}_K]\in{\mathbb R}^{M\times K}$ and ${\bf \Sigma}=\diag(\sigma_1, \cdots, \sigma_k, \cdots, \sigma_K)\in{\mathbb R}^{K\times K}$, where $\{{\bf v}_k, \sigma_k\}$ are the eigenvector-eigenvalue pairs~\cite{GH:icml2011}. Figure~\ref{fig:twobitscomparision} shows the two-bit partitioning on a synthetic data with nonlinear structure using different hashing methods, including spectral hashing, exact graph hashing, and anchor graph hashing. Note that since spectral hashing computes two smoothest pseudo graph Laplacian eigenfunctions instead of performing real spectral embedding, it can not handle such type of nonlinear data structures. The exact graph hashing method first constructs an exact neighborhood graph, e.g., $k$NN graph, and then performs partitioning with spectral techniques to solve the optimization problem in Eq.\ref{SHproperties}. The anchor graph hashing archives a good separation (by the first bit) of the nonlinear manifold and balancing partitioning, even performs better than the exact graph hashing, which loses the property of balancing partitioning for the second bit. The anchor graph hashing approach was recently further improved by leveraging a discrete optimization technique to directly solve binary hash codes without any relaxation~\cite{Liu:2014NIPS}.

\subsection{Angular Quantization Based Hashing}
\begin{figure}[t]
	\begin{center}
		\centerline{\includegraphics[width=0.62\columnwidth]{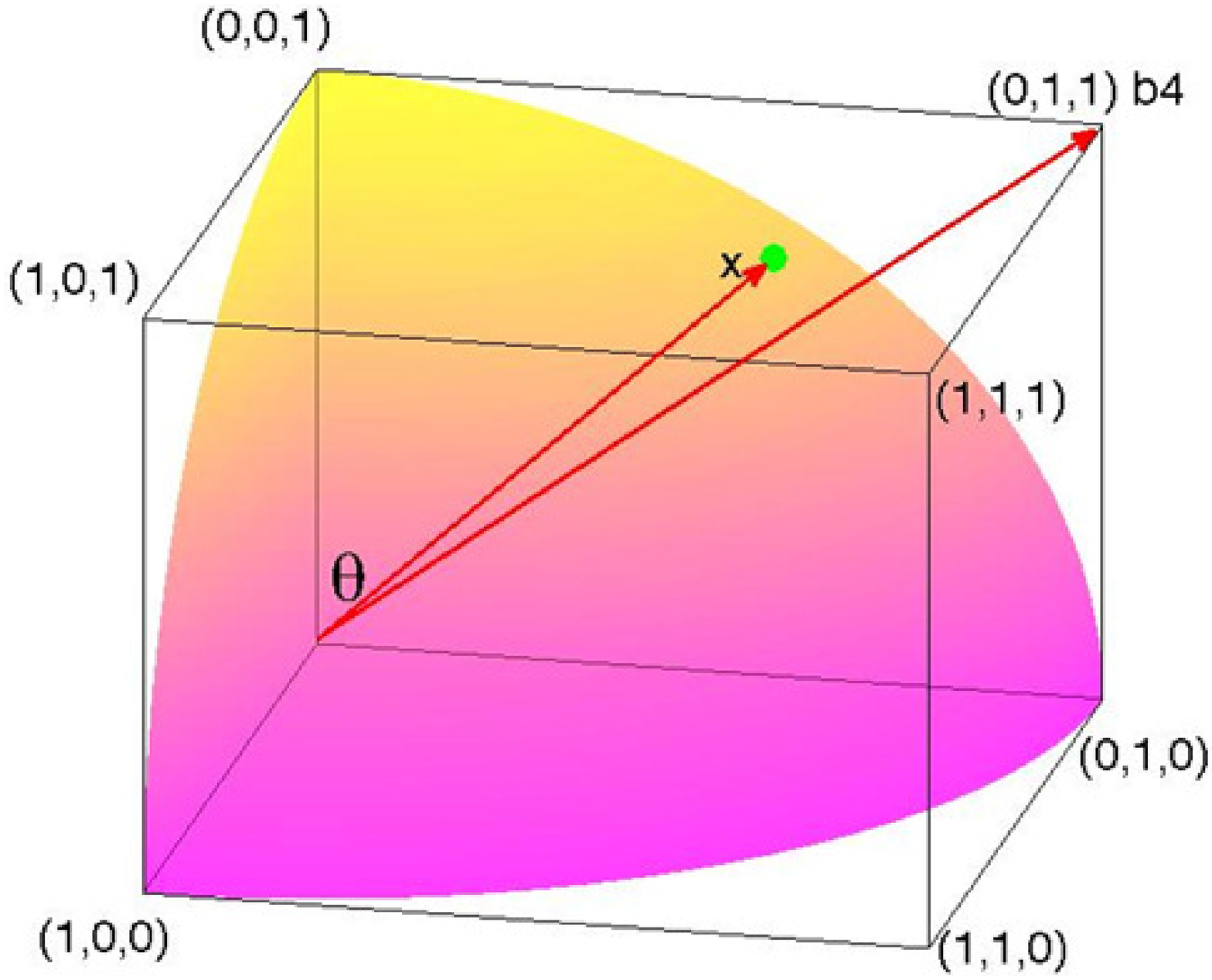}}
		\caption{Illustration of angular quantization based hashing method~\protect\cite{AQ:2012NIPS}. The binary code of a data point ${\bf x}$ is assigned as the nearest binary vertex in the hypercube, which is ${\bf b}_4=[0\;\;1\;\;1]^\top$ in the illustrated example~\protect\cite{AQ:2012NIPS}.}
		\label{fig:AGHmethod}
	\end{center}
\end{figure}

Since similarity is often measured by the cosine of the angle between pairs of samples, angular quantization is thus proposed to map non-negative feature vectors onto a vertex of the binary hypercube with the smallest angle~\cite{AQ:2012NIPS}. In such a setting, the vertices of the hypercube is treated as quantization landmarks that grow exponentially with the data dimensionality $D$. As shown in Figure~\ref{fig:AGHmethod}, the nearest binary vertex ${\bf b}$ in a hypercube to the data point ${\bf x}$ is given by
\begin{eqnarray}
\label{AQHobjective}
&&{\bf b}^\ast=\arg\max_{\bf b}\frac{{\bf b}^\top {\bf x}}{\|{\bf b}\|_2}\\ \nonumber
&&\textrm{subject to:}~~~~~{\bf b}\in \{0, 1\}^K,
\end{eqnarray}
Although it is an integer programming problem, its global maximum can be found with a complexity of ${\cal O}(D\log D)$. The optimal binary vertices will be used as the binary hash codes for data points as ${\bf y}={\bf b}^\ast$. Based on this angular quantization framework, a data-dependent extension is designed to learn a rotation matrix ${\bf R}\in{\mathbb R}^{D\times D}$ to align the projected data ${\bf R}^\top {\bf x}$ to the binary vertices without changing the similarity between point pairs. The objective is formulated as the following
\begin{eqnarray}
\label{AQHroatationobjective}
&&({\bf b}_i^\ast, {\bf R}^\ast)=\arg\max_{{\bf b}_i, {\bf R}}\sum_i\frac{{\bf b}_i^\top}{\|{\bf b}_i\|_2}{\bf R}^\top{\bf x}_i\\ \nonumber
&&\textrm{subject to:}~~~~~{\bf b}\in \{0, 1\}^K\\ \nonumber
&&~~~~~~~~~~~~~~~~~~~{\bf R}^\top{\bf R}={\bf I}_{D\times D}
\end{eqnarray}
Note that the above formulation still generates a $D$-bit binary code for each data point, while compact codes are often desired in many real-world applications~\cite{SSH:2012PAMI}. To generate a $K$=bit code, a projection matrix ${\bf S}\in{\mathbb R}^{D\times K}$ with orthogonal columns can be used to replace the rotation matrix $\bf R$ in the above objective with additional normalization, as discussed in~\cite{AQ:2012NIPS}. Finally, the optimal binary codes and the projection/rotation matrix are learned using an alternating optimization scheme.

\subsection{Binary Reconstructive Embedding}
Instead of using data-independent random projections as in {\it LSH} or principal components as in {\it SH}, Kulis and
Darrell~\cite{Kulis:2009NIPS} proposed data-dependent and
bit-correlated hash functions as:
\begin{eqnarray}\label{BREhash}
h_k({\bf x})=\sgn\left (\sum_{q=1}^s {\bf W}_{kq}\kappa({\bf x}_{kq},{\bf x}) \right)
\end{eqnarray}
The sample set $\{{\bf x}_{kq}\}, q=1,\cdots, s$ is the training data for learning hash function $h_k$ and $\kappa(\cdot)$ is a kernel function, and ${\bf W}$ is a weight matrix.

Based on the above formulation, a method called Binary Reconstructive Embedding ({\it BRE}) was designed to minimize a cost function measuring the difference between the metric and reconstructed distance in hamming space. The Euclidean metric $d_{\cal M}$ and the binary reconstruction distance $d_{\cal R}$ are defined as:
\begin{eqnarray}\label{BREdistance}
&&d_{\cal M}({\bf x}_i,{\bf x}_j)=\frac{1}{2}\|{\bf x}_i-{\bf x}_j\|^2\\ \nonumber
&&d_{\cal R}({\bf x}_i,{\bf x}_j)=\frac{1}{K}\sum_{k=1}^K\left (h_k({\bf x}_i)-h_k({\bf x}_j)\right )^2
\end{eqnarray}
The objective is to minimize the following reconstruction error to derive the optimal {\bf W}:
\begin{eqnarray}\label{BREobjective}
{\bf W}^\ast=\arg\min_{\bf W}\sum_{({
\bf x}_i, {\bf x}_j)\in {\cal N}}\left [d_{\cal M}({\bf x}_i,{\bf x}_j)-d_{\cal R}({\bf x}_i,{\bf x}_j)\right ]^2,
\end{eqnarray}
where the set of sample pairs $\cal N$ is the training data. Optimizing the above objective function is difficult due to the non-differentiability of $\sgn(\cdot)$ function. Instead, a coordinate-descent algorithm was applied to iteratively update the hash functions to a local optimum. This hashing method can be easily extended to a supervised scenario by setting pairs with same labels to have zero distance and pairs with different labels to have a large distance. However, since the binary reconstruction distance $d_{\cal R}$ is bounded in $[0, 1]$ while the metric distance $d_{\cal M}$ has no upper bound, the minimization problem in Eq.~(\ref{BREobjective}) is only meaningful when input data is appropriately normalized. In practice, the original data point ${\bf x}$ is often mapped to a hypersphere with unit length so that $0\le d_{\cal M}\le 1$. This normalization removes the scale of data points, which is often not negligible for practical applications of nearest neighbor search. In addition, Hamming distance based objective is hard to optimize due to its nonconvex and nonsmooth properties. Hence, Liu et al. proposed to utilize the equivalence between code inner products and the Hamming distances to design supervised and kernel-based hash functions~\cite{SKH:cvpr2012}. The objective is to ensure the inner product of hash codes consistent with the given pairwise supervision. Such a strategy of optimizing the hash code inner product in KSH rather than the Hamming distance like what's done in BRE pays off nicely and leads to major performance gains in similarity-based retrieval consistently confirmed in extensive experiments reported in~\cite{SKH:cvpr2012} and recent studies~\cite{SparseHash14}.

\begin{figure}[t]
\begin{center}
\centerline{\includegraphics[width=0.96\columnwidth]{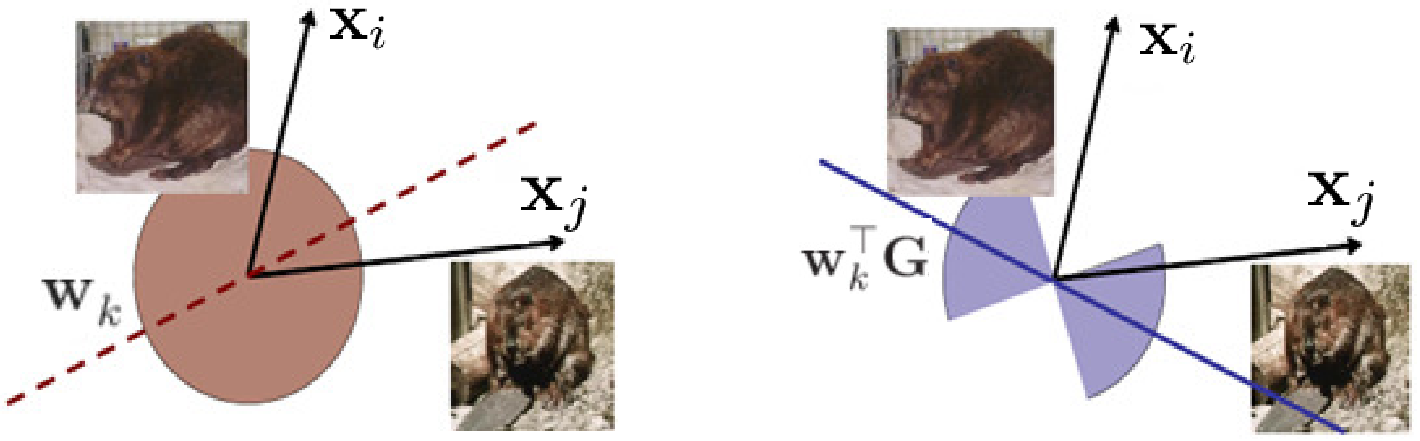}}
\caption{Illustration of the hashing method based on metric learning. The left shows the partitioning using standard LSH method and the right shows the partitioning of the metric learning based LSH method (modified the original figure in~\protect\cite{Kulis:2009PAMI}).}
\label{fig:MLhashingmethod}
\end{center}
\end{figure}

\subsection{Metric Learning based Hashing}
The key idea for metric learning based hashing method is to learn a parameterized Mahalanobis metric using pairwise label information. Such learned metrics are then employed to the standard random projection based hash functions~\cite{Kulis:2009PAMI}. The goal is to preserve the pairwise relationship in the binary code space, where similar data pairs are more likely to collide in the same hash buck and dissimilar pairs are less likely to share the same hash codes, as illustrated in Figure~\ref{fig:MLhashingmethod}.

The parameterized inner product is defined as $$sim({\bf x}_i,{\bf x}_j)={\bf x}_i^\top{\bf M}{\bf x}_j,$$ where ${\bf M}$ is a positive-definite $d\times d$ matrix to be learned from the labeled data. Note that this similarity measure corresponds to the parameterized squared Mahalanobis distance $d_{\bf M}$. Assume that ${\bf M}$ can be factorized as ${\bf M}~=~{\bf G}^\top {\bf G}$. Then the parameterized squared Mahalanobis distance can be written as
\begin{eqnarray}
\label{metriclearning}
d_{\bf M}({\bf x}_i,{\bf x}_j)&=&({\bf x}_i-{\bf x}_j)^\top {\bf M}({\bf x}_i-{\bf x}_j)\\ \nonumber
&=&({\bf G}{\bf x}_i-{\bf G}{\bf x}_i)^\top({\bf G}{\bf x}_i-{\bf G}{\bf x}_i).
\end{eqnarray}
Based on the above equation, the distance $d_{\bf M}({\bf x}_i,{\bf x}_j)$ can be interpreted as the Euclidian distance between the projected data points ${\bf G}{\bf x}_i$ and ${\bf G}{\bf x}_j$. Note that the matrix ${\bf M}$ can be learned through various metric learning method such as information-theoretic metric learning~\cite{Davis:2007ICML}. To accommodate the learned distance metric, the randomized hash function is given as
\begin{eqnarray}
\label{metriclearninghashfunction}
h_k({\bf x}) = \sgn({\bf w}_k{\bf G}^\top {\bf x}).
\end{eqnarray}
It is easy to see that the above hash function generates the hash codes which preserve the parameterized similarity measure in the Hamming space. Figure~\ref{fig:MLhashingmethod} demonstrates the difference between standard random projection based LSH and the metric learning based LSH, where it is easy to see that the learned metric help assign the same hash bit to the similar sample pairs. Accordingly, the collision probability is given as
\begin{eqnarray}
\label{MLSHcollision}
{\textrm Pr}\left [h_k({\bf x}_i)=h_k({\bf x}_i)\right ]=1-\frac{1}{\pi}\cos^{-1}\frac{{\bf x}_i^\top{\bf G}^\top{\bf G}{\bf x}_j}{\|{\bf G}{\bf x}_i\|\|{\bf G}{\bf x}_j\|}
\end{eqnarray}
Realizing that the pairwise constraints often come to be available incrementally, Jain et al exploit an efficient online locality-sensitive hashing with gradually learned distance metrics~\cite{Jain:2008NIPS}.

\begin{figure}[t]
 \centering \setlength{\fboxrule}{0.2pt}
 \subfigure[ ]
 {\label{fig:threeprojectionbasehashing:a}
 \fbox{\includegraphics[width=0.28\linewidth, height=0.18\linewidth]{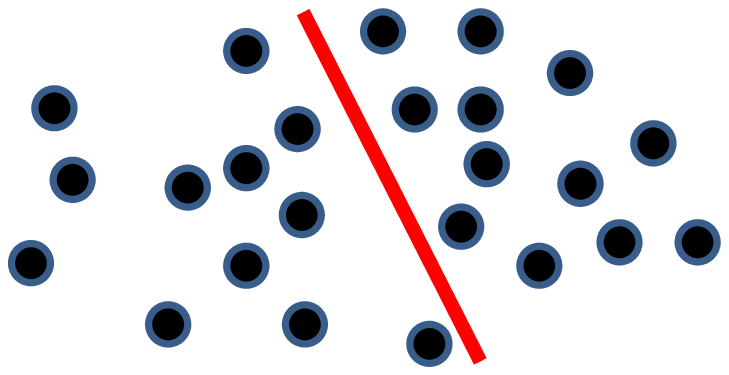}}}
  \subfigure[ ]
 {\label{fig:threeprojectionbasehashing:b}
 \fbox{\includegraphics[width=0.28\linewidth, height=0.18\linewidth]{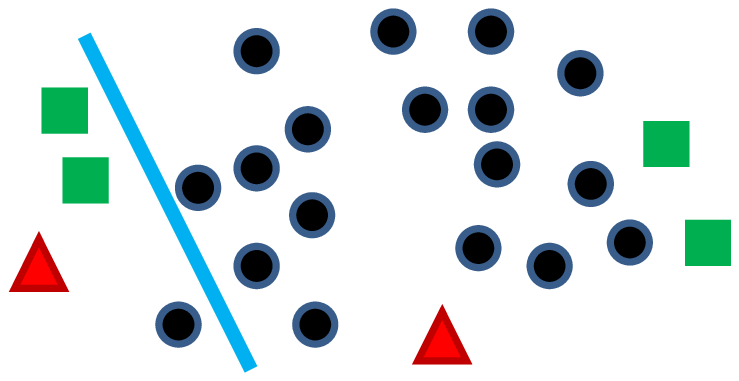}}}
  \subfigure[ ]
 {\label{fig:threeprojectionbasehashing:c}
 \fbox{\includegraphics[width=0.28\linewidth, height=0.18\linewidth]{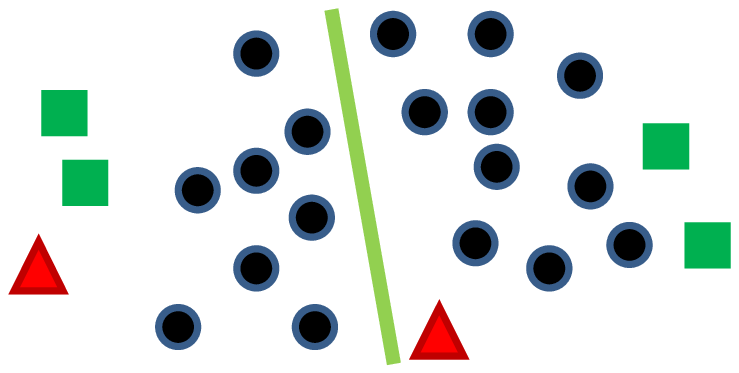}}}
 \caption{Illustration of one-bit partitioning of different linear projection based hashing methods: a) unsupervised hashing; b) supervised hashing; and c) semi-supervised hashing. The similar point pairs are indicated in the green rectangle shape and the dissimilar point pairs are with a red triangle shape.}
\label{fig:threeprojectionbasehashing}
\end{figure}

\subsection{Semi-Supervised Hashing}
Supervised hashing techniques have been shown to be superior than unsupervised approaches since they leverage the supervision information to design task-specific hash codes. However, for a typical setting of large scale problem, the human annotation process is often costly and the labels can be noisy and sparse, which could easily lead to overfitting.

Considering a small set of pairswise labels and a large amount of unlabled data, semi-supervised hashing aims in designing hash functions with minimum empirical loss while maintaining maximum entropy over the entire dataset. Specifically, assume the pairwise labels are given as two type of sets ${\cal M}$ and $\cal C$. A pair of data point $({\bf x}_i, {\bf x}_j)\in {\cal M}$ indicates that ${\bf x}_i$ and ${\bf x}_j$ are similar and $({\bf x}_i, {\bf x}_j)\in {\cal C}$ means that ${\bf x}_i$ and ${\bf x}_j$. Hence, the empirical accuracy on the labeled data for a family of hash functions ${\bf H}=[h_1, \cdots, h_K]$ is given as
\begin{eqnarray}
\label{SRHEMPRICALCost}
\!\!\!\!\!\!J({\bf H})\!\!\!=\!\!\!\sum\limits_k\!\!\!\left [\!\! \sum\limits_{({\bf x}_i, {\bf x}_j)\in {\cal M}}\!\!\! \!\!\!  h_k({\bf x}_i)h_k({\bf x}_j)-\!\!\!\!\!\!\!\!\! \sum\limits_{({\bf x}_i, {\bf x}_j)\in {\cal C}}\!\!\!\!\!\!\!\!\! h_k({\bf x}_i)h_k({\bf x}_j) \!\!\right ].
\end{eqnarray}
If define a matrix ${\bf S} \in {\mathbb R}^{l\times l}$ incorporating the pairwise labeled information from ${\bf X}_l$ as:
\begin{eqnarray}
\label{Eq constainedmatrix}
S_{ij}=\left \{ \begin{array} {r@{\quad: \quad}l}
1 & ({\bf x}_i, {\bf x}_j)\in {\cal M}\\ -1& ({\bf x}_i, {\bf x}_j)\in {\cal C}\\
0 & otherwise.
\end{array}\right. ,
\end{eqnarray}
The above empirical accuracy can be written in a compact matrix form after dropping off the $\sgn(\cdot)$ function
\begin{eqnarray}\label{matrixformcostj}
 J({\bf H}) =\frac{1}{2}{\bf tr}\left ({\bf W}^\top {\bf X}_l~{\bf S}~{\bf W}^\top {\bf X}_l^\top \right ).
\end{eqnarray}
However, only considering the empirical accuracy during the design of the hash function can lead to undesired results. As illustrated in Figure~\ref{fig:threeprojectionbasehashing:b}, although such a hash bit partitions the data with zero error over the pairwise labeled data, it results in imbalanced separation of the unlabeled data, thus being less informative. Therefore, an information theoretic regularization is suggested to maximize the entropy of each hash bit. After relaxation, the final objective is formed as
\begin{eqnarray}
\label{singlebitrevisedcost}\nonumber
{\bf W}^\ast=\arg\max_{\bf W}\frac{1}{2}{\bf tr}\left ({\bf W}^\top {\bf X}_l{\bf S}{\bf X}_l^\top {\bf W}\right )+\frac{\eta}{2} {\bf tr}\left ({\bf W}^\top {\bf X} {\bf X}^\top {\bf W}\right ),\\
\end{eqnarray}
where the first part represents the empirical accuracy and the second component encourages partitioning along large variance projections. The coefficient $\eta$ weighs the contribution from these two components. The above objective can be solved using various optimization strategies, resulting in orthogonal or correlated binary codes, as described in~\cite{SSH:2012PAMI}\cite{SPLH:icml2010}. Figure~\ref{fig:threeprojectionbasehashing} illustrates the comparison of one-bit linear partition using different learning paradigms, where the semi-supervised method tends to produce balanced yet accurate data separation. Finally, Xu et al. employs similar semi-supervised formulation to sequentially learn multiple complementary hash tables to further improve the performance~\cite{Xu:2011ICCV}.

\subsection{Column Generation Hashing}
Beyond pairwise relationship, complex supervision like ranking triplets and ranking lists has been exploited to learn hash functions with the property of ranking preserving. In many real applications such as image retrieval and recommendation system, it is often easier to receive the relative comparison instead of instance-wise or pari-wise labels. For a general claim, such relative comparison information is given in a triplet form. Formally, a set of triplets are represented as: $${\cal E}=\{({\bf q}_i, {\bf x}_i^+, {\bf x}_i^-)|sim({\bf q}_i, {\bf x}_i^+)>sim({\bf q}_i, {\bf x}_i^-)\},$$ where the the function $sim(\cdot)$ could be an unknown similarity measure. Hence, the triplet $({\bf q}_i, {\bf x}_i^+, {\bf x}_i^-)$ indicates that the sample point ${\bf x}_i^+$ is more semantically similar or closer to a query point ${\bf q}$ than the point ${\bf x}_i^-$, as demonstrated in Figure~\ref{fig:differentsupervision:b}.

As one of the representative methods falling into this category, column generation hashing explores the large-margin framework to leverage such type of proximity comparison information to design weighted hash functions~\cite{CG:2013ICML}. In particular, the relative comparison information $sim({\bf q}_i, {\bf x}_i^+)>sim({\bf q}_i, {\bf x}_i^-)$ will be preserved in a weighted Hamming space as $d_{{\cal WH}}({\bf q}_i, {\bf x}_i^+)<d_{{\cal WH}}({\bf q}_i, {\bf x}_i^-)$, where $d_{\cal WH}$ is the weighted Hamming distance as defined in Eq.~\ref{weightedhammingdis}. To impose a large-margin, the constraint $d_{{\cal WH}}({\bf q}_i, {\bf x}_i^+)<d_{{\cal WH}}({\bf q}_i, {\bf x}_i^-)$ should be satisfied as well as possible. Thus, a typical large-margin objective with $\ell_1$ norm regularization can be formulated as
\begin{eqnarray}
\label{Eq CGHobjective}
&&\arg\min_{{\bf w}, \xi}\sum_{i=1}^{|{\cal E}|}\xi_i+C\|{\bf w}\|_1 \\ \nonumber
&&\textrm{subject to:}~~~{\bf w}\succeq {\bf 0}, {\boldsymbol \xi}\succeq {\bf 0}; \\ \nonumber
&&~~~~~~~~~~~~~~~~d_{{\cal WH}}({\bf q}_i, {\bf x}_i^-)-d_{{\cal WH}}({\bf q}_i, {\bf x}_i^-)\ge 1-\xi_i, \forall i,
\end{eqnarray}
where ${\bf w}$ is the random projections for computing the hash codes. To solve the above optimization problem, the authors proposed using column generation technique to learn the hash function and the associated bit weights iteratively. For each iteration, the best hash function is generated and the weight vector is updated accordingly. In addition, different loss functions with other regularization terms such as $\ell_\infty$ are also suggested as alternatives in the above formulation.

\begin{figure}[t]
\begin{center}
\centerline{\includegraphics[width=1\columnwidth, height=0.56\columnwidth]{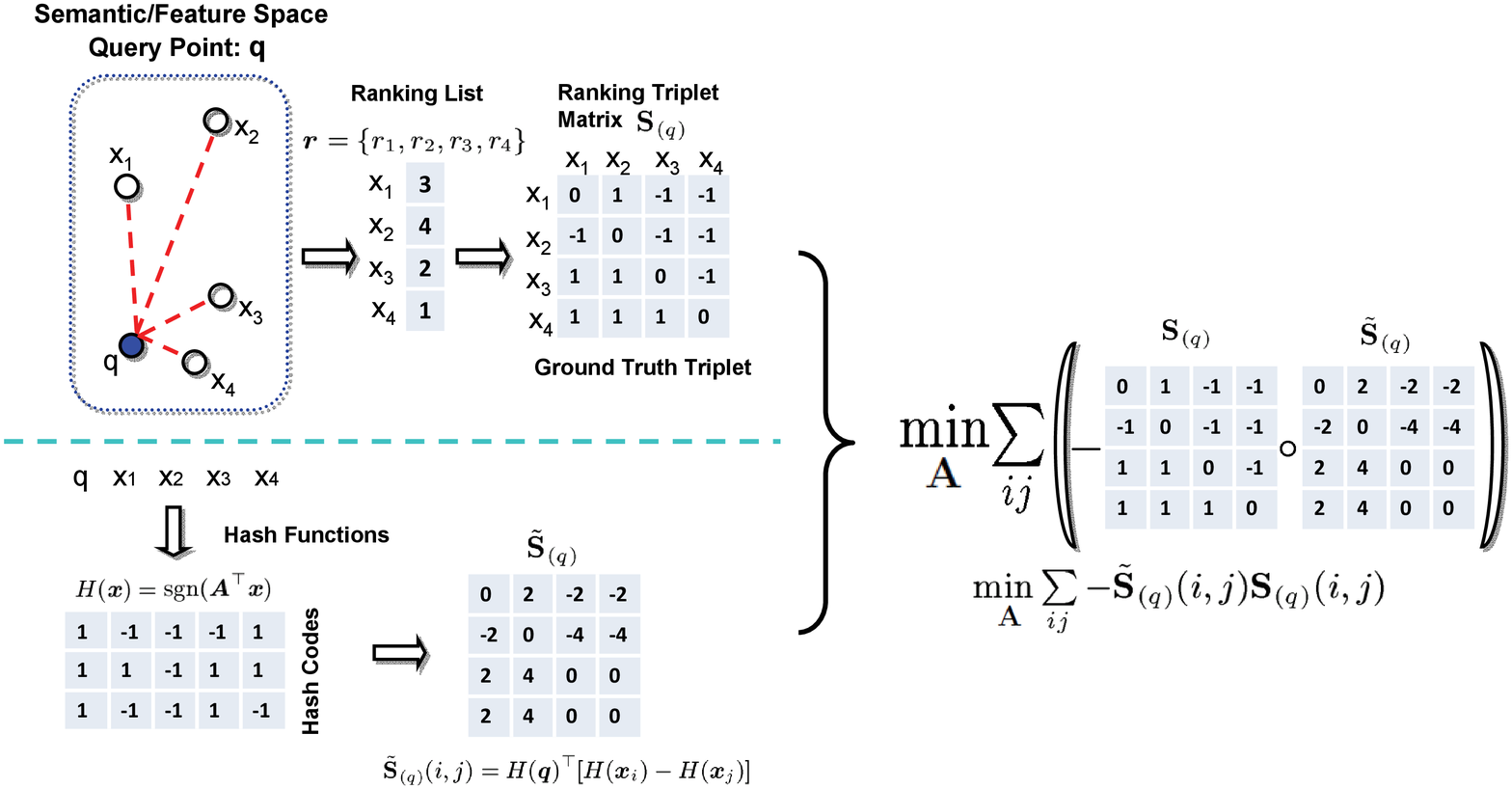}}
\caption{The conceptual diagram of the {\em Rank Supervised Hashing} method. The top left component demonstrates the procedure of deriving ground truth ranking list ${\bf r}$ using the semantic relevance or feature similarity/distance, and then converting it to a {\em triplet matrix} ${\bf S}_{(q)}$ for a given query $q$. The bottom left component describes the estimation of relaxed ranking {\em triplet matrix} ${\tilde {\bf S}}_{(q)}$ from the binary hash codes. The right component shows the objective of minimizing the inconsistency between the two ranking {\em triplet matrices}.}
\label{fig intuition}
\end{center}
\end{figure}

\subsection{Ranking Supervised Hashing}
Different from other methods that explore the triplet relationship~\cite{CG:2013ICML}\cite{Wang:2013ACMMM}\cite{Norouzi:NIPS2012}, the ranking supervised hashing method attempt to preserve the ranking order of a set of database points corresponding to the query point~\cite{ListwiseH:2013ICCV}. Assume that the training dataset ${\cal X}=\{{\bf x}_n\}$ has $N$ points with ${\bf x}_n~\in~{\mathbb R}^D$. In addition, a query set is given as ${\cal Q}~=~\{{\bf q}_m\}$, and ${\bf q}_m~\in~{\mathbb R}^D, m=1, \cdots, M$. For any specific query point ${\bf q}_m$, we can derive a ranking list over $\cal X$, which can be written as a vector as $r({\bf q}_m,{\cal X})=(r_1^m, \cdots, r_n^m, \cdots, r_N^m)$. Each element $r_n^m$ falls into the integer range $[1, N]$ and no two elements share the same value for the exact ranking case. If $r_i^m<r_j^m$ ($i,j=1, \cdots, N$), it indicates sample ${\bf x}_i$ has higher rank than ${\bf x}_j$, which means ${\bf x}_i$ is more relevant or similar to ${\bf q}_m$ than ${\bf x}_j$. To represent such a discrete ranking list, a ranking {\em triplet matrix} ${\bf S}\in{\mathbb R}^{N\times N}$ is defined as
\begin{eqnarray}
\label{Eq triplet}
S({\bf q}_m; {\bf x}_i, {\bf x}_j)=\left \{ \begin{array} {r@{\quad: \quad}l}
1 & r_i^q<r_j^q\\ -1& r_i^q>r_j^q\\
0 & r_i^q=r_j^q.
\end{array}\right.
\end{eqnarray}
Hence for a set of query points ${\cal Q}=\{{\bf q}_m\}$, we can derive a {\em triplet tensor}, i.e., a set of triplet matrices $${\bf S}=\{{\bf S}_{({\bf q}_m)}\}~\in~{\mathbb R}^{M\times N \times N}.$$ In particular, the element of the triplet tensor is defined as
${\bf S}_{mij}={\bf S}_{({\bf q}_m)}(i,j)=S({\bf q}_m; {\bf x}_i, {\bf x}_j)$, $m=1, \cdots, M$, $i, j=1, \cdots, N$. The objective is to preserve the ranking lists in the mapped Hamming space. In other words, if $S({\bf q}_m; {\bf x}_i, {\bf x}_j)=1$, we tend to ensure $d_{\cal H}({\bf q}_m, {\bf x}_i)~<~d_{\cal H}({\bf q}_m, {\bf x}_i)$, otherwise $d_{\cal H}({\bf q}_m, {\bf x}_i)~>~d_{\cal H}({\bf q}_m, {\bf x}_i)$. Assume the hash code has the value as $\{-1, 1\}$, such ranking order is equivalent to the similarity measurement using the inner products of the binary codes, i.e., $$d_{\cal H}({\bf q}_m, {\bf x}_i)\!\!<\!\!d_{\cal H}({\bf q}_m, {\bf x}_i) \!\!\Leftrightarrow\!\! H({\bf q}_m)^\top H({\bf x}_i)\!\!>\!\!H({\bf q}_m)^\top H({\bf x}_j).$$
Then the empirical loss function ${L}_{\cal H}$ over the ranking list can be represented as 
$${L}_{\cal H}=-\sum_{m}\sum_{i,j}H({\bf q}_m)^\top\left [H({\bf x}_i)-H({\bf x}_j)\right ]S_{mij}.$$
Assume we utilize linear hash functions, the final objective is formed as the following constrained quadratic problem
\begin{eqnarray}
\label{Eq finaloptproblemwithconstraints}
{\bf W}^{\ast}&=&\arg\max_{\bf W}{L}_H=\arg\max_{\bf W}\tr({\bf W}{\bf W}^\top{\bf B}) \\ \nonumber
 &\text{s.t.}&{\bf W}^\top{\bf W}={\bf I},
\end{eqnarray}
where the constant matrix $\bf B$ is computed as ${\bf B}=\sum_m{\bf p}_m{\bf q}_m^\top$ with ${\bf p}_m=\sum_{i,j}\left [{\bf x}_i-{\bf x}_j\right ]S_{mij}$. The orthogonality constraint is utilized to minimize the redundancy between different hash bits. Intuitively, the above formulation is to preserve the ranking list in the Hamming space, as shown in the conceptual diagram in Figure~\ref{fig intuition}. The augmented Lagrangian multiplier method was introduced to derive feasible solutions for the above constrained problem, as discussed in~~\cite{ListwiseH:2013ICCV}.

\subsection{Circulant Binary Embedding}
Realizing that most of the current hashing techniques rely on linear projections, which could suffer from very high computational and storage costs for high-dimensional data, circulant binary embedding was recently developed to handle such a challenge using the circulant projection~\cite{YU:icml2014}. Briefly, given a vector ${\bf r}=\{r_0, \cdots, r_{d-1}\}$, we can generate its corresponding circulant matrix ${\bf R}=circ({\bf r})$~\cite{Gray:2006toeplitz}. Therefore, the binary embedding with the circulant projection is defined as:
\begin{eqnarray}
\label{Eq circulantprojection}
h({\bf x})=\sgn({\bf Rx})=\sgn(circ({\bf r})\cdot{\bf x}).
\end{eqnarray}
Since the circulant projection $circ({\bf r}){\bf x}$ is equivalent to circular convolution ${\bf r}\ocoasterisk{\bf x}$, the computation of linear projection can be eventually realized using fast Fourier transform as  
\begin{eqnarray}
\label{Eq circulantprojectionFFT}
circ({\bf r}){\bf x}= {\bf r}\ocoasterisk{\bf x}={\cal F}^{-1}\left ({\cal F}({\bf r})\circ {\cal F}({\bf x}) \right ).
\end{eqnarray}
Thus, the time complexity is reduced from $d^2$ to $d\log{d}$. Finally, one could randomly select the circulant vector ${\bf r}$ or design specific ones using supervised learning methods.

\section{Deep Learning for Hashing}
\label{sec:dnnhash}
During the past decade (since around 2006), \textit{Deep Learning}~\cite{DL15}, also known as \textit{Deep Neural Networks}, has drawn increasing attention and research efforts in a variety of artificial intelligence areas including speech recognition, computer vision, machine learning, text mining, \textit{etc}. Since one main purpose of deep learning is to learn robust and powerful feature representations for complex data, it is very natural to leverage deep learning for exploring compact hash codes which can be regarded as binary representations of data. In this section, we briefly introduce several recently proposed hashing methods that employ deep learning. In Table~\ref{tab:deeplearninghash}, we compare eight deep learning based hashing methods in terms of four key characteristics that can be used to differentiate the approaches.

The earliest work in deep learning based hashing may be \textit{Semantic Hashing}~\cite{SemanticHash09}. This method builds a deep generative model to discover hidden binary units (\textit{i.e.}, latent topic features) which can model input text data (\textit{i.e.}, word-count vectors). Such a deep model is made as a stack of \textit{Restricted Boltzmann Machines} (RBMs)~\cite{Hinton:2006Science}. After learning a multi-layer RBM through pre-training and fine-tuning on a collection of documents, the hash code of any document is acquired by simply thresholding the output of the deepest layer. Such hash codes provided by the deep RBM were shown to preserve semantically similar relationships among input documents into the code space, in which each hash code (or hash key) is used as a memory address to locate corresponding documents. In this way, semantically similar documents are mapped to adjacent memory addresses, thereby enabling efficient search via hash table lookup. To enhance the performance of deep RBMs, a supervised version was proposed in~\cite{Torralba:2008CVPR}, which borrows the idea of nonlinear Neighbourhood Component Analysis (NCA) embedding~\cite{NNCA07}. The supervised information stems from given neighbor/nonneighbor relationships between training examples. Then, the objective function of NCA is optimized on top of a deep RBM, making the deep RBM yield discriminative hash codes. Note that supervised deep RBMs can be applied to broad data domains other than text data. In \cite{Torralba:2008CVPR}, supervised deep RBMs using a Gaussian distribution to model visible units in the first layer were successfully applied to handle massive image data.

\begin{table*}[t]
	\centering
	\caption{The characteristics of eight recently proposed deep learning based hashing methods. }
	\begin{tabular}{|c|c|c|c|c|}
		\hline 
		\multicolumn{1}{|c|}{\textbf{\small Deep Learning based}}  &\multicolumn{1}{c|}{\textbf{\small Data}}  &\multicolumn{1}{c|}{\textbf{\small Learning}} &\multicolumn{1}{c|}{\textbf{\small Learning}} &\multicolumn{1}{c|}{\textbf{\small Hierarchy of}} \\
		\textbf{\small Hashing Methods} &\textbf{\small Domain} &\textbf{\small Paradigm} &\textbf{\small Features?} &\textbf{\small Deep Neural Networks} \\
		\hline\hline \small
		Semantic Hashing \cite{SemanticHash09}                   &\small text           &\small unsupervised &\small no   &\small 4       \\ \hline
		\small Restricted Boltzmann Machine \cite{Torralba:2008CVPR}                &\small text and image &\small supervised   &\small no   &\small 4 and 5 \\ \hline
		\small Tailored Feed-Forward Neural Network \cite{SparseHash14} &\small text and image &\small supervised   &\small no   &\small 6       \\ \hline
		\small Deep Hashing \cite{DeepHash15}                           &\small image          &\small unsupervised &\small no   &\small 3       \\ \hline
		\small Supervised Deep Hashing \cite{DeepHash15}                &\small image          &\small supervised   &\small no   &\small 3       \\ \hline
		\small Convolutional Neural Network Hashing \cite{CNNHash14}    &\small image          &\small supervised   &\small yes  &\small 5       \\ \hline
		\small Deep Semantic Ranking Hashing \cite{RankHash15}          &\small image          &\small supervised   &\small yes  &\small 8       \\ \hline
		\small Deep Neural Network Hashing \cite{DNNHash15}             &\small image          &\small supervised   &\small yes  &\small 10      \\ \hline
	\end{tabular}
	\label{tab:deeplearninghash}
\end{table*}

A recent work named \textit{Sparse Similarity-Preserving Hashing}~\cite{SparseHash14} tried to address the low recall issue pertaining to relatively long hash codes, which affect most of previous hashing techniques. The idea is enforcing sparsity into the hash codes to be learned from training examples with pairwise supervised information, that is, similar and dissimilar pairs of examples (also known as side information in the machine learning literature). The relaxed hash functions, actually nonlinear embedding functions, are learned by training a \textit{Tailored Feed-Forward Neural Network}. Within this architecture, two ISTA-type networks \cite{SparseCode10} that share the same set of parameters and conduct fast approximations of sparse coding are coupled in the training phase. Since each output of the neural network is continuous albeit sparse, a hyperbolic tangent function is applied to the output followed by a thresholding operation, leading to the final binary hash code. In \cite{SparseHash14}, an extension to hashing multimodal data, \textit{e.g.}, web images with textual tags, was also presented.

Another work named \textit{Deep Hashing} \cite{DeepHash15} developed a deep neural network to learn a multiple hierarchical nonlinear transformation which maps original images to compact binary hash codes and hence supports large-scale image retrieval with the learned binary image representation. The deep hashing model is established under three constraints which are imposed on the top layer of the deep neural network: 1) the reconstruction error between an original real-valued image feature vector and the resulting binary code is minimized,
2) each bit of binary codes has a balance, and 3) all bits are independent of each other. Similar constraints have been adopted in prior unsupervised hashing or binary coding methods such as Iterative Quantization (ITQ) \cite{ITQ13}. A supervised version called \textit{Supervised Deep Hashing}\footnote{It is essentially semi-supervised as abundant unlabeled examples are used for training the deep neural network.} was also presented in \cite{DeepHash15}, where a discriminative term incorporating pairwise supervised information is added to the objective function of the deep hashing model. The authors of \cite{DeepHash15} showed the superiority of the supervised deep hashing model over its unsupervised counterpart. Both of them produce hash codes through thresholding the output of the top layer in the neural network, where all activation functions are hyperbolic tangent functions.

It is worthwhile to point out that the above methods, including Sparse Similarity-Preserving Hashing, Deep Hashing and Supervised Deep Hashing, did not include a pre-training stage during the training of the deep neural networks. Instead, the hash codes are learned from scratch using a set of training data. However, the absence of pre-training may make the generated hash codes less effective. Specifically, the Sparse Similarity-Preserving Hashing method is found to be inferior to the state-of-the-art supervised hashing method Kernel-Based Supervised Hashing (KSH)~\cite{SKH:cvpr2012} in terms of search accuracy on some image datasets~\cite{SparseHash14}; the Deep Hashing method and its supervised version are slightly better than ITQ and its supervised version CCA+ITQ, respectively~\cite{ITQ13}\cite{DeepHash15}. Note that KSH, ITQ and CCA+ITQ exploit relatively shallow learning frameworks.

Almost all existing hashing techniques including the aforementioned ones relying on deep neural networks take a vector of hand-crafted visual features extracted from an image as input. Therefore, the quality of produced hash codes heavily depends on the quality of hand-crafted features. To remove this barrier, a recent method called \textit{Convolutional Neural Network Hashing} \cite{CNNHash14} was developed to integrate image feature learning and hash value learning into a joint learning model. Given pairwise supervised information, this model consists of a stage of learning approximate hash codes and a stage of training a deep \textit{Convolutional Neural Network} (CNN)~\cite{CNN12} that outputs continuous hash values. Such hash values can be generated by activation functions like sigmoid, hyperbolic tangent or softmax, and then quantized into binary hash codes through appropriate thresholding. Thanks to the power of CNNs, the joint model is capable of simultaneously learning image features and hash values, directly working on raw image pixels. The deployed CNN is composed of three convolution-pooling layers that involve rectified linear activation, max pooling, and local contrast normalization, a standard fully-connected layer, and an output layer with softmax activation functions.

Also based on CNNs, a latest method called as \textit{Deep Semantic Ranking Hashing} \cite{RankHash15} was presented to learn hash values such that multilevel semantic similarities among multi-labeled images are preserved. Like the Convolutional Neural Network Hashing method, this method takes image pixels as input and trains a deep CNN, by which image feature representations and hash values are jointly learned. The deployed CNN consists of five convolution-pooling layers, two fully-connected layers, and a hash layer (\textit{i.e.}, output layer). The key hash layer is connected to both fully-connected layers and in the function expression as $$h({\bf x})=2\sigma\big(\mathbf{w}^\top [f_1({\bf x});f_2({\bf x})]\big)-1,$$ in which ${\bf x}$ represents an input image, $h({\bf x})$ represents the vector of hash values for image ${\bf x}$, $f_1({\bf x})$ and $f_2({\bf x})$ respectively denote the feature representations from the outputs of the first and second fully-connected layers, $\mathbf{w}$ is the weight vector, and $\sigma()$ is the logistic function. The Deep Semantic Ranking Hashing method leverages listwise supervised information to train the CNN, which stems from a collection of image triplets that encode the multilevel similarities, \textit{i.e.}, the first image in each triplet is more similar to the second one than the third one. The hash code of image ${\bf x}$ is finally obtained by thresholding the output $h({\bf x})$ of the hash layer at zero. 

The above Convolutional Neural Network Hashing method~\cite{CNNHash14} requires separately learning approximate hash codes to guide the subsequent learning of image representation and finer hash values.A latest method called \textit{Deep Neural Network Hashing}~\cite{DNNHash15} goes beyond, in which the image representation and hash values are learned in one stage so that representation learning and hash learning are tightly coupled to benefit each other. Similar to the Deep Semantic Ranking Hashing method~\cite{RankHash15}, the Deep Neural Network Hashing method incorporates listwise supervised information to train a deep CNN, giving rise to a currently deepest architecture for supervised hashing. The pipeline of the deep hashing architecture includes three building blocks: 1) a triplet of images (the first image is more similar to the second one than the third one) which are fed to the CNN, and upon which a triplet ranking loss is designed to characterize the listwise supervised information; 2) a shared sub-network with a stack of eight convolution layers to generate the intermediate image features; 3) a divide-and-encode module to divide the intermediate image features into multiple channels, each of which is encoded into a single hash bit. Within the divide-and-encode module, there are one fully-connected layer and one hash layer. The former uses sigmoid activation, while the latter uses a piecewise thresholding scheme to produce a nearly discrete hash values. Eventually, the hash code of any image is yielded by thresholding the output of the hash layer at 0.5. In~\cite{DNNHash15}, the Deep Neural Network Hashing method was shown to surpass the Convolutional Neural Network Hashing method as well as several shallow learning based supervised hashing methods in terms of image search accuracy. 

Last, we a few observations are worth mentioning deep learning based hashing methods introduced in this section.
\begin{enumerate}
  \item The majority of those methods did not report the time of hash code generation.   In real-world search scenarios, the speed for generating hashes should be substantially fast. There might be concern about the hashing speed of those deep neural network driven approaches, especially the approaches involving image feature learning, which may take much longer time to hash an image compared to shallow learning driven approaches like ITQ and KSH.    
  
  \item Instead of employing deep neural networks to seek hash codes, another interesting problem is to design a proper hashing technique to accelerate deep neural network training or save memory space. The latest work \cite{NNHash15} presented a hashing trick named \textit{HashedNets}, which shrinks the storage costs of neural networks significantly while mostly preserving the generalization performance in image classification tasks.
\end{enumerate}

\section{Advanced Methods and Related Applications}
\label{sec:advances}
In this section, we further extend the survey scope to cover a few more advanced hashing methods that are developed for specific settings and applications, such as point-to-hyperplane hashing, subspace hashing, and multimodality hashing.

\begin{figure}[t]
\begin{center}
\centerline{\includegraphics[width=0.96\columnwidth]{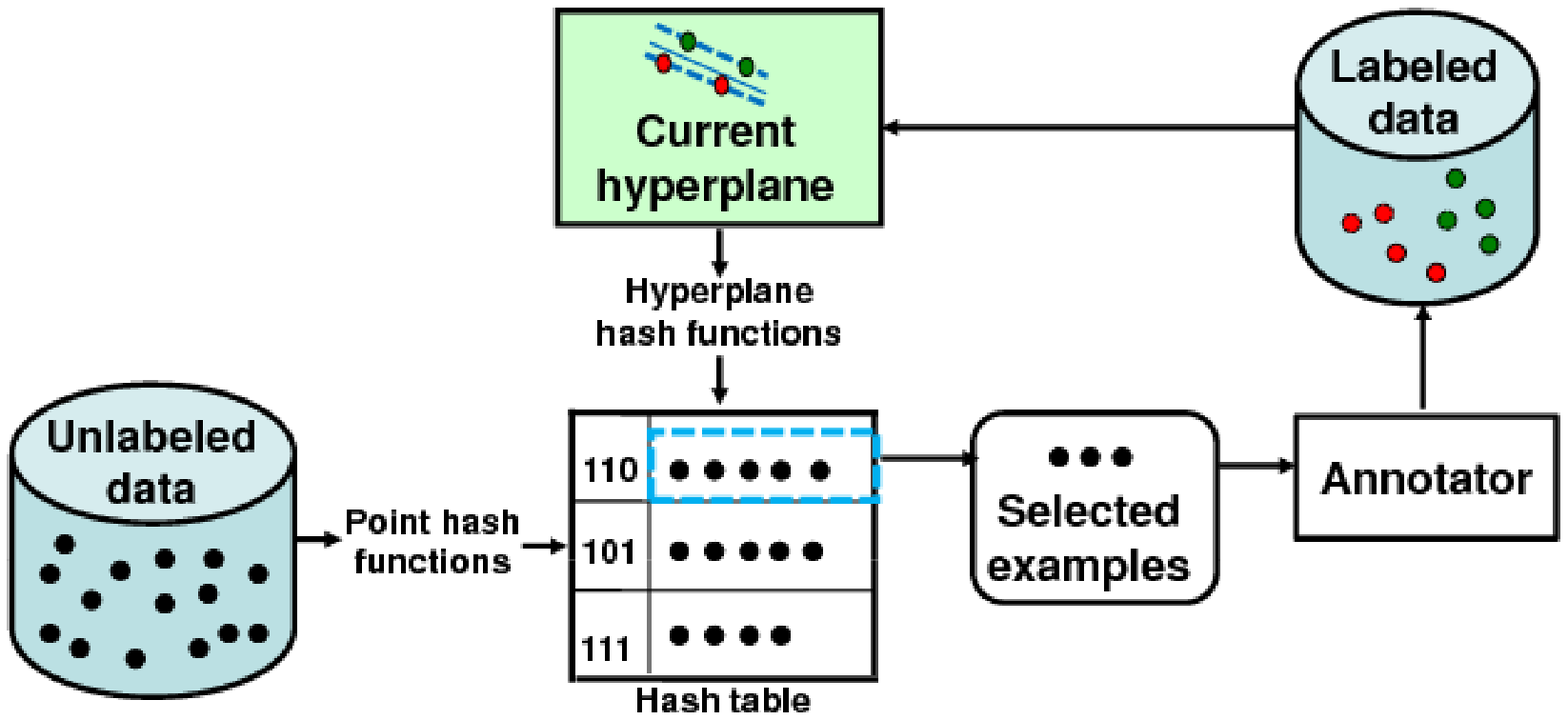}}
\caption{Active learning framework with a hashing-based fast query selection strategy~\protect\footnotemark.}
\label{fig:ActiveLearningHash}
\end{center}
\end{figure}
\footnotetext{The illustration figure is from http://vision.cs.utexas.edu/projects/activehash/}

\subsection{Hyperplane Hashing}
\begin{figure}[t]
\begin{center}
\centerline{\includegraphics[width=0.92\columnwidth]{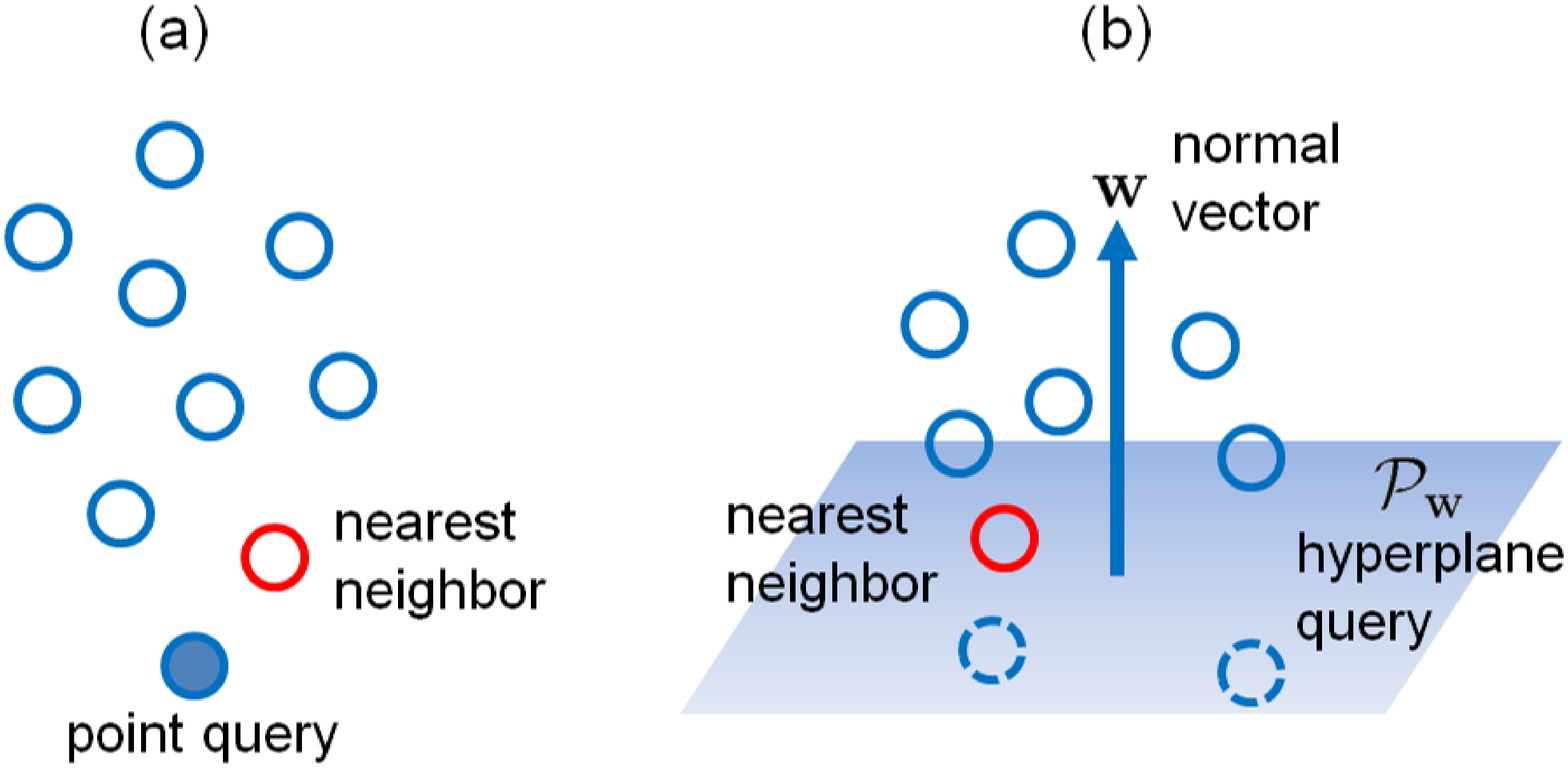}}
\caption{Two distinct nearest neighbor search problems. (a) Point-to-point search,
  the blue solid circle represents a point query and the red circle represents
  the found nearest neighbor point. (b) Point-to-hyperplane search,
  the blue plane denotes a hyperplane query $\mathcal{P}_{\mathbf{w}}$ with $\mathbf{w}$ being its normal vector,
  and the red circle denotes the found nearest neighbor point.}
\label{fig:hyperplane_problems}
\end{center}
\end{figure}

Distinct from the previously surveyed conventional hashing techniques all of which address the problem of fast point-to-point nearest neighbor search (see Figure~\ref{fig:hyperplane_problems}(a)), a new scenario ``point-to-hyperplane" hashing emerges to tackle fast point-to-hyperplane nearest neighbor search (see Figure~\ref{fig:hyperplane_problems}(b)), where the query is a hyperplane instead of a data point. Such a new scenario requires hashing the hyperplane query to near database points, which is difficult to accomplish because point-to-hyperplane distances are quite different from routine point-to-point distances in terms of the computation mechanism. Despite the bulk of research on point-to-point hashing, this special hashing paradigm is rarely touched. For convenience, we call point-to-hyperplane hashing as \textit{Hyperplane Hashing}.

Hyperplane hashing is actually fairly important for many machine learning applications such as large-scale active learning with SVMs~\cite{Vijayanarasimhan:2013PAMI}. In SVM-based active learning \cite{Tong:2001JMLR}, the well proven sample selection strategy is to search in the unlabeled sample pool to identify the sample closest to the current hyperplane decision boundary, thus providing the most useful information for improving the learning model. When making such active learning scalable to gigantic databases, exhaustive search for the point nearest to the hyperplane is not efficient for the online sample selection requirement. Hence, novel hashing methods that can principally handle hyperplane queries are called for. A conceptual diagram using hyperplane hashing to scale up active learning process is demonstrated in Figure~\ref{fig:ActiveLearningHash}.

\begin{figure}[t]
\begin{center}
\centerline{\includegraphics[width=0.92\columnwidth]{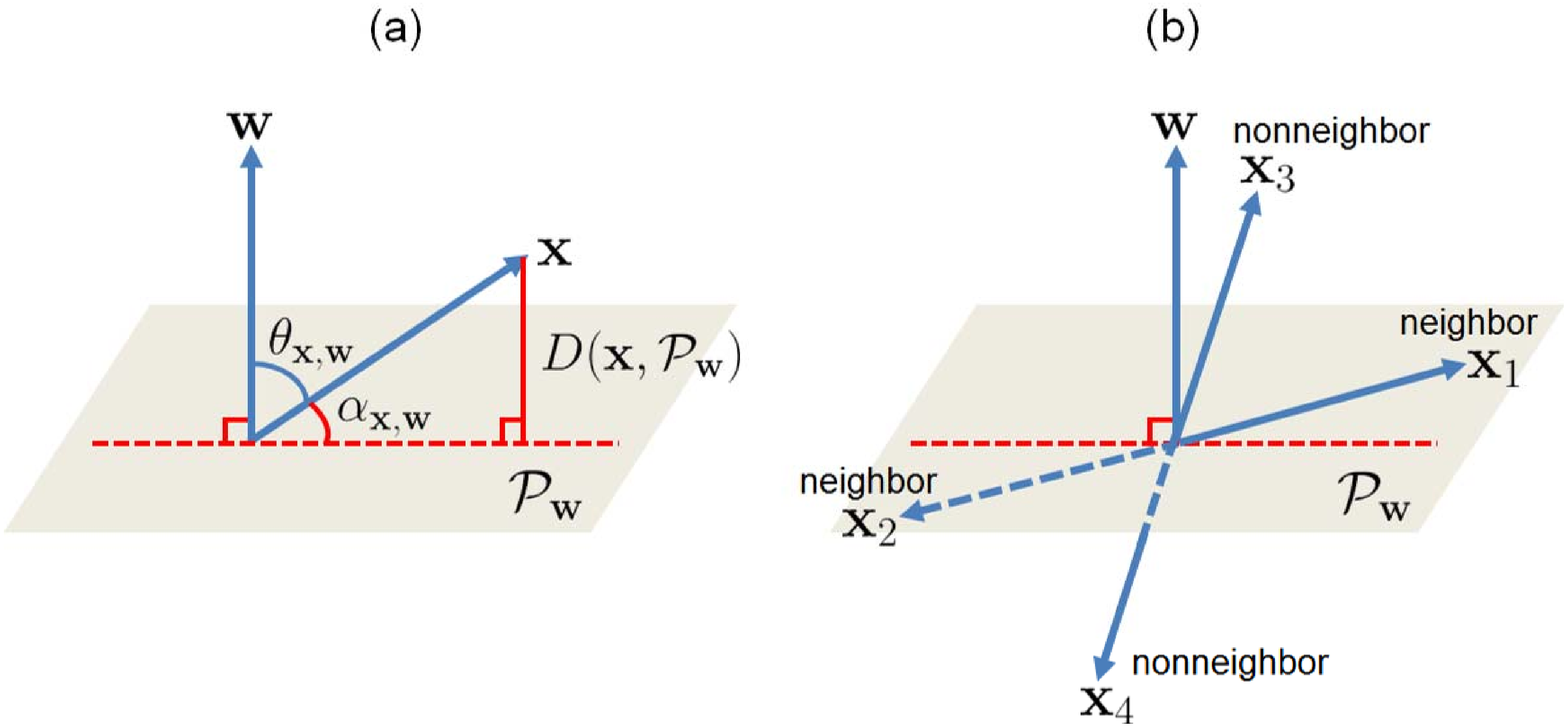}}
\caption{The hyperplane hashing problem. (a) Point-to-hyperplane distance
  $D(\mathbf{x},\mathcal{P}_{\mathbf{w}})$ and point-to-hyperplane angle $\alpha_{\mathbf{x},\mathbf{w}}$.
  (b) Neighbors ($\bm{x}_1,\bm{x}_2$) and nonneighbors ($\bm{x}_3,\bm{x}_4$) of the hyperplane query $\mathcal{P}_{\mathbf{w}}$,
  and the ideal neighbors are the points $\perp\mathbf{w}$.}
\label{fig:hyperplane_toy}
\end{center}
\end{figure}

We demonstrate the geometric relationship between a data point $\mathbf{x}$ and a hyperplane $\mathcal{P}_{\mathbf{w}}$ with the vector normal as $\mathbf{w}$ in Figure~\ref{fig:hyperplane_toy}(a). Given a hyperplane query $\mathcal{P}_{\mathbf{w}}$ and a set of points
$\mathcal{X}$, the target nearest neighbor is $$\mathbf{x}^*=\arg\min_{\mathbf{x}\in\mathcal{X}}D(\mathbf{x},\mathcal{P}_{\mathbf{w}}),$$
where $D(\mathbf{x},\mathcal{P}_{\mathbf{w}})=\frac{|\mathbf{w}^\top\mathbf{x}|}{\|\mathbf{w}\|}$ is the \textit{point-to-hyperplane distance}. The existing hyperplane hashing methods~\cite{Jain:2010NIPS}\cite{Liu:2012ICML} all attempt to minimize a slightly modified ``distance" $\frac{|\mathbf{w}^\top\mathbf{x}|}{\|\mathbf{w}\|\|\mathbf{x}\|}$, i.e., the sine of the \textit{point-to-hyperplane angle}
$\alpha_{\mathbf{x},\mathbf{w}}=\left|\theta_{\mathbf{x},\mathbf{w}}-\frac{\pi}{2}\right|$. Note that $\theta_{\mathbf{x},\mathbf{w}}\in[0,\pi]$ is the angle between $\mathbf{x}$ and $\mathbf{w}$. The angle measure $\alpha_{\mathbf{x},\mathbf{w}}\in[0,\pi/2]$ between a database point and a hyperplane query turns out to be reflected into the design of hash functions.

As shown in Figure~\ref{fig:hyperplane_toy}(b), the goal of hyperplane hashing is to hash a hyperplane query $\mathcal{P}_{\mathbf{w}}$ and the desired neighbors (e.g., $\mathbf{x}_1,\mathbf{x}_2$) with narrow $\alpha_{\mathbf{x},\mathbf{w}}$ into the same or nearby hash buckets, meanwhile avoiding to return the undesired nonneighbors (e.g., $\mathbf{x}_3,\mathbf{x}_4$) with wide $\alpha_{\mathbf{x},\mathbf{w}}$. Because $\alpha_{\mathbf{x},\mathbf{w}}=\left|\theta_{\mathbf{x},\mathbf{w}}-\frac{\pi}{2}\right|$, the point-to-hyperplane search problem can be equivalently transformed to a specific point-to-point search problem where the query is the hyperplane normal $\mathbf{w}$ and the desired nearest neighbor to the raw query $\mathcal{P}_{\mathbf{w}}$ is the one whose angle $\theta_{\mathbf{x},\mathbf{w}}$ from $\mathbf{w}$ is closest to $\pi/2$, i.e., most closely perpendicular to $\mathbf{w}$ (we write ``perpendicular to $\mathbf{w}$" as $\perp\mathbf{w}$ for brevity). This is very different from traditional point-to-point nearest neighbor search which returns the most similar point to the query point. In the following, several existing hyperplane hashing methods will be briefly discussed

Jain \textit{et al.} \cite{Liu:2012ICML} devised two different families of randomized hash functions to attack the hyperplane hashing problem.
The first one is \textit{Angle-Hyperplane Hash} (AH-Hash) $\mathcal{A}$, of which one instance function is
\begin{flalign} \label{eq:ahash}
&~h^{\mathcal{A}}(\mathbf{z})=\nonumber \\
&\left\{\begin{matrix}
[\textrm{sgn}(\mathbf{u}^\top\mathbf{z}),\textrm{sgn}(\mathbf{v}^\top\mathbf{z})],\mathbf{z}~\textrm{is a database point}  \\
[\textrm{sgn}(\mathbf{u}^\top\mathbf{z}),\textrm{sgn}(-\mathbf{v}^\top\mathbf{z})],~~~\mathbf{z}~\textrm{is a hyperplane normal}
\end{matrix}\right.
\end{flalign}
where $\mathbf{z}\in\mathbb{R}^d$ represents an input vector, and $\mathbf{u}$ and $\mathbf{v}$ are both drawn independently from a standard $d$-variate Gaussian, i.e., $\mathbf{u},\mathbf{v}\thicksim\mathcal{N}(0,I_{d\times d})$. Note that $h^{\mathcal{A}}$ is a two-bit hash function which leads to the probability of collision for a hyperplane normal $\mathbf{w}$ and a database point $\mathbf{x}$:
\begin{flalign} \label{eq:aprob}
\mathbf{Pr}\left[h^{\mathcal{A}}(\mathbf{w})=h^{\mathcal{A}}(\mathbf{x})\right]=\frac{1}{4}-\frac{\alpha^2_{\mathbf{x},\mathbf{w}}}{\pi^2}.
\end{flalign}
This probability monotonically decreases as the point-to-hyperplane angle $\alpha_{\mathbf{x},\mathbf{w}}$ increases, ensuring angle-sensitive
hashing.

The second family proposed by Jain \textit{et al.} is \textit{Embedding-Hyperplane Hash} (EH-Hash) function family $\mathcal{E}$ of which one instance is
\begin{flalign} \label{eq:ehash}
  h^{\mathcal{E}}(\mathbf{z})=\left\{\begin{matrix}
      \textrm{sgn}\left(\mathbf{U}^\top\mathbf{V}(\mathbf{z}\mathbf{z}^\top)\right),\mathbf{z}~\textrm{is a database point}  \\
      \textrm{sgn}\left(-\mathbf{U}^\top\mathbf{V}(\mathbf{z}\mathbf{z}^\top)\right),~~\,\mathbf{z}~\textrm{is
        a hyperplane normal}
\end{matrix}\right.
\end{flalign}
where $\mathbf{V}(A)$ returns the vectorial concatenation of matrix $A$, and $\mathbf{U}\thicksim\mathcal{N}(0,I_{d^2\times d^2})$. The EH hash function $h^{\mathcal{E}}$ yields hash bits on an embedded space $\mathbb{R}^{d^2}$ resulting from vectorizing rank-one matrices $\mathbf{z}\mathbf{z}^\top$ and $-\mathbf{z}\mathbf{z}^\top$. Compared with $h^{\mathcal{A}}$, $h^{\mathcal{E}}$ gives a higher probability of collision:
\begin{flalign} \label{eq:eprob}
\mathbf{Pr}\left[h^{\mathcal{E}}(\mathbf{w})=h^{\mathcal{E}}(\mathbf{x})\right]=\frac{\cos^{-1}\sin^2(\alpha_{\mathbf{x},\mathbf{w}})}{\pi},
\end{flalign}
which also bears the angle-sensitive hashing property. However, it is much more expensive to compute than AH-Hash.

\begin{figure}
\centering
\hskip -0.2in
\begin{minipage}{.24\textwidth}
\scalebox{0.22}{\includegraphics{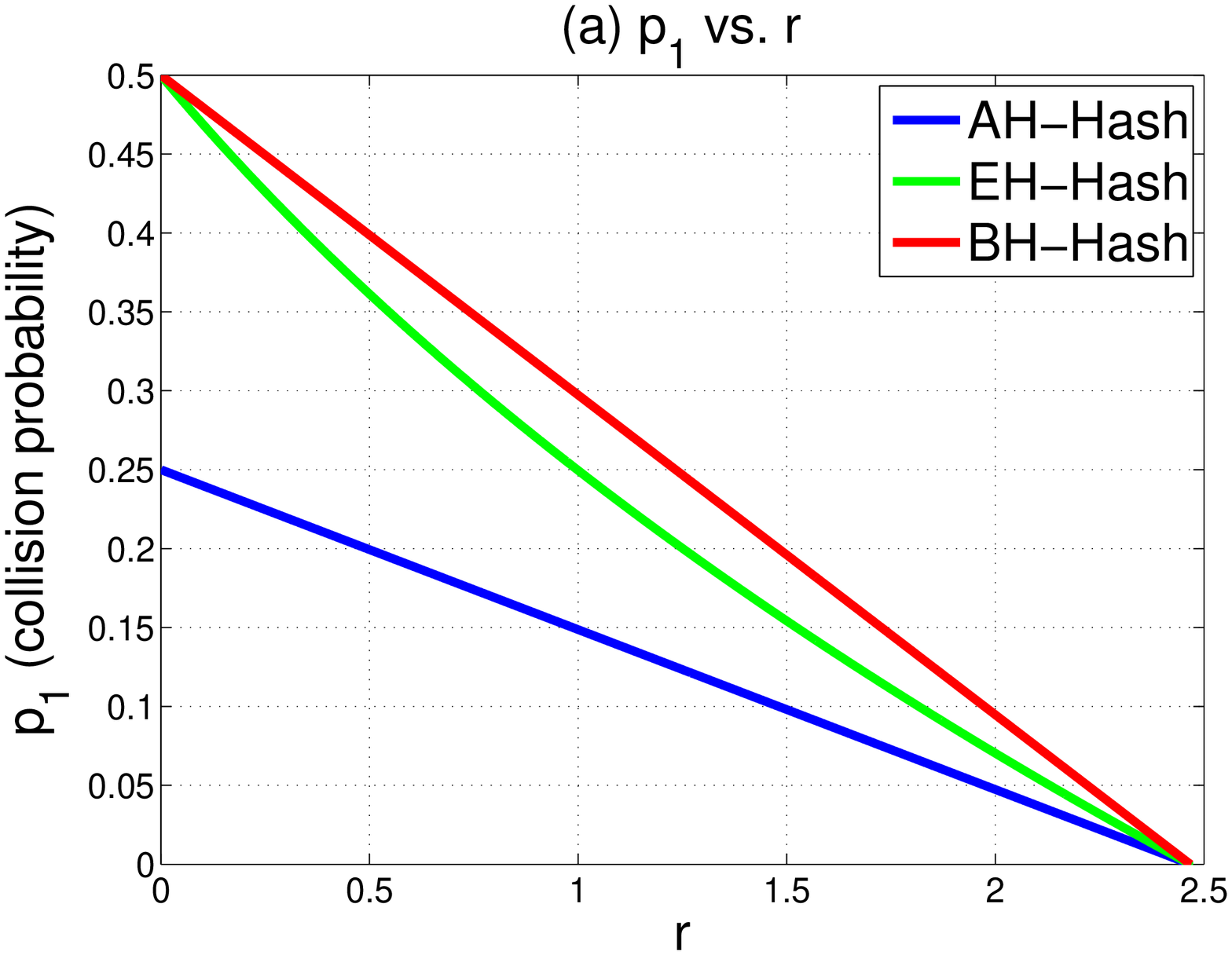}}
\end{minipage}
\hskip 0.1in
\caption{Comparison of the collision probabilities of the three randomized hyperplane hashing schemes using $p_1$ (probability of collision) vs. $r$ (squared point-to-hyperplane angle)}
  \label{fig:theory_compare}
\vskip -0.05in
\end{figure}

More recently, Liu \textit{et al.} \cite{Liu:2012ICML} designed a randomized function family with bilinear
\textit{Bilinear-Hyperplane Hash} (BH-Hash) as:
\begin{flalign}\label{eq:bfamily}
  \mathcal{B}=\left\{h^{\mathcal{B}}(\mathbf{z})=\mathrm{sgn}(\mathbf{u}^\top\mathbf{z}\mathbf{z}^\top\mathbf{v}),~
    \textrm{i.i.d.~}\mathbf{u},\mathbf{v}\thicksim\mathcal{N}(0,I_{d\times
      d})\right\}.
\end{flalign}
As a core finding, Liu \textit{et al.} proved in \cite{Liu:2012ICML} that
the probability of collision for a hyperplane query $\mathcal{P}_{\mathbf{w}}$ and
a database point $\mathbf{x}$ under $h^{\mathcal{B}}$ is
\begin{flalign} \label{eq:bprob}
\mathbf{Pr}\left[h^{\mathcal{B}}(\mathcal{P}_{\mathbf{w}})=h^{\mathcal{B}}(\mathbf{x})\right]
=\frac{1}{2}-\frac{2\alpha^2_{\mathbf{x},\mathbf{w}}}{\pi^2}.
\end{flalign}
Specifically, $h^{\mathcal{B}}(\mathcal{P}_{\mathbf{w}})$ is prescribed to be $-h^{\mathcal{B}}(\mathbf{w})$.
Eq.~(\ref{eq:bprob}) endows $h^{\mathcal{B}}$ with the angle-sensitive hashing property. It is important to find that the collision probability given by the BH hash function $h^{\mathcal{B}}$ is always twice of the collision probability by the AH hash function $h^{\mathcal{A}}$, and also greater than the collision probability by the EH hash function $h^{\mathcal{E}}$. As illustrated in Figure~\ref{fig:theory_compare}, for any fixed $r$, BH-Hash accomplishes the highest probability of collision, which indicates that the BH-Hash has a better angle-sensitive property.

In terms of the formulation, the bilinear hash function $h^{\mathcal{B}}$ is correlated with yet different from the linear hash functions $h^{\mathcal{A}}$ and $h^{\mathcal{E}}$. (1) $h^{\mathcal{B}}$ produces a single hash bit which is the product of the two hash bits produced by $h^{\mathcal{A}}$. (2) $h^{\mathcal{B}}$ may be a rank-one special case of $h^{\mathcal{E}}$ in algebra if we write $\mathbf{u}^\top\mathbf{z}\mathbf{z}^\top\mathbf{v}=\mathrm{tr}(\mathbf{z}\mathbf{z}^\top\mathbf{v}\mathbf{u}^\top)$ and $\mathbf{U}^\top\mathbf{V}(\bm{z}\bm{z}^\top)=\mathrm{tr}(\bm{z}\bm{z}^\top\mathbf{U})$.
(3) $h^{\mathcal{B}}$ appears in a universal form, while both $h^{\mathcal{A}}$ and $h^{\mathcal{E}}$ treat a query and a database item in a distinct manner. The computation time of $h^{\mathcal{B}}$ is $\Theta(2d)$ which is the same as that of $h^{\mathcal{A}}$ and one order of magnitude faster than $\Theta(2d^2)$ of $h^{\mathcal{E}}$. Liu \textit{et al.} further improved the performance of $h^{\mathcal{B}}$ through learning the bilinear projection directions $\mathbf{u},\mathbf{v}$ in $h^{\mathcal{B}}$ from the data. Gong et al. extended the bilinear formulation to the conventional point-to-point hashing scheme through designing compact binary codes for high-dimensional visual descriptors~\cite{GongBL:cvpr2013}.  

\subsection{Subspace Hashing}

Beyond the aforementioned conventional hashing which tackles searching in a database of vectors, \textit{subspace hashing}~\cite{ANSS:2011PAMI}, which has been rarely explored in the literature, attempts to efficiently search through a large database of subspaces. Subspace representation is very common in many computer vision, pattern recognition, 
and statistical learning problems, such as subspace representations of image patches, image sets, video clips, \textit{etc}. For example, face images of the same subject with fixed poses but different illuminations are often assumed to reside near linear subspaces. A common use scenario is to use a single face image to find the subspace (and the corresponding subject ID) closest to the query image~\cite{SubspaceH:2013ICCV}. Given a query in the form of vector or subspace, searching for a nearest subspace in a subspace database is frequently encountered in a variety of practical applications including example-based image synthesis, scene classification, speaker recognition, face recognition, and motion-based action recognition~\cite{SubspaceH:2013ICCV}.

However, hashing and searching for subspaces are both different from the schemes used in traditional vector hashing and the latest hyperplane hashing. \cite{ANSS:2011PAMI} presented a general framework to the problem of \textit{Approximate Nearest Subspace} (ANS) search, which uniformly deals with the cases that query is a vector or subspace, query and database elements are subspaces of fixed dimension, query and database elements are subspaces of different dimension, and database elements are subspaces of varying dimension. The critical technique exploited by \cite{ANSS:2011PAMI} is two-step: 1) a simple mapping that maps both query and database elements to ``points" in a new vector space, and 2) doing approximate nearest neighbor search using conventional vector hashing algorithms in the new space. Consequently, the main contribution of \cite{ANSS:2011PAMI} is reducing the difficult subspace hashing problem to a regular vector hashing task. \cite{ANSS:2011PAMI} used LSH for the vector hashing task. While simple, the hashing technique (mapping + LSH) of \cite{ANSS:2011PAMI} perhaps suffers from the high dimensionality of the constructed new vector space.

More recently, \cite{SubspaceH:2013ICCV} exclusively addressed the
point-to-subspace query where query is a vector and database items are subspaces of arbitrary dimension. \cite{SubspaceH:2013ICCV} proposed a rigorously faster hashing technique than that of \cite{ANSS:2011PAMI}. 
Its hash function can hash $D$-dimensional vectors ($D$ is the ambient dimension of the query) or $D\times r$-dimensional subspaces ($r$ is arbitrary) in a linear time complexity $O(D)$, which is computationally more efficient than the hash functions devised in \cite{ANSS:2011PAMI}. \cite{SubspaceH:2013ICCV} further proved the search time under the $O(D)$ hashes
to be sublinear in the database size.  

Based on the nice finding of~\cite{SubspaceH:2013ICCV}, we would like to achieve
faster hashing for the subspace-to-subspace query by means of crafted novel hash functions to handle subspaces in varying dimension. Both theoretical and practical explorations in this direction will be beneficial to the hashing area.

\subsection{MultiModality Hashing}
Note that the majority of the hash learning methods are designed for constructing the Hamming embedding for a single modality or representation. Some recent advanced methods are proposed to design the hash functions for more complex settings, such as that the data are represented by multimodal features or the data are formed in a heterogeneous way~\cite{SSLH:2012ECCV}. Such type of hashing methods are closely related to the applications in social network, whether multimodality and heterogeneity are often observed. Below we survey several representative methods that are proposed recently.

Realizing that data items like webpage can be described from multiple information sources, composing hashing was recently proposed to design hashing schme using several information sources~\cite{Zhang:2011SIGIR}. Besides the intuitive way of concatenating multiple features to derive hash functions, the author also presented an iterative weighting scheme and formulated convex combination of multiple features. The objective is to ensure the consistency between the semantic similarity and Hamming similarity of the data. Finally, a joint optimization strategy is employed to learn the importance of individual type of features and the hash functions. Co-regularized hashing was proposed to investigate the hashing learning across multiparity data in a supervised setting, where similar and dissimilar pairs of intra-modality points are given as supervision information~\cite{Zhen:2012NIPS}. One of such a typical setting is to index images and the text jointly to preserve the semantic relations between image and text. The authors formulate their objective as a boosted co-regularization framework with the cost component as a weighted sum of the intra-modality and inter-modality loss. The learning process of the hash functions is performed via a boosting procedure to that the bias introduced by previous hash function can be sequentially minimized. Dual-view hashing attempts to derive a hidden common Hamming embedding of data from two views, while maintaining the predictability of the binary codes~\cite{PDVH:2013ICML}. A probabilistic model called multimodal latent binary embedding was recently presented to derive binary latent factors in a common Hamming space for indexing multimodal data~\cite{PMH:2012KDD}. Other closely related hashing methods include the design of multiple feature hashing for near-duplicate duplicate detection~\cite{Song:2011ACMMM}, submodular hashing for video indexing~\cite{Cao:2012ACMMM}, and Probabilistic Attributed Hashing for integrating low-level features and semantic attributes~\cite{PAH:2015AAAI}.

\subsection{Applications with Hashing}
Indexing massive multimedia data, such as images and video, are the natural applications for learning based hashing. Especially, due to the well-known semantic gap, supervised and semi-supervised hashing methods have been extensively studied for image search and retrieval~\cite{Jain:2008CVPR}\cite{SSH:cvpr2010}\cite{Kulis:2009ICCV}\cite{Grauman:2013LBH}\cite{Grauman:2013ESSI}\cite{Kong:2012SIGIR}, mobile product search\cite{MPS:2012CVPR}. Other closely related computer vision applications include image patch matching~\cite{SBHJSD:2011ICCV}, image classification~\cite{GongBL:cvpr2013}, face recognition~\cite{FRH:cvpr2010}\cite{OBH:2015IJCAI}, pose estimation~\cite{Shakhnarovich:2003ICCV}, object tracking~\cite{HforVT:cvpr2013}, and duplicate detection~\cite{Song:2011ACMMM}\cite{PQH:2012ECCV}\cite{PMH:2010ECCV}\cite{NDID:2008BMVC}\cite{Manku:2007WWW}. In addition, this emerging hash learning framework can be exploited for some general machine learning and data mining tasks, including cross-modality data fusion~\cite{DF:cvpr2010}, large scale optimization~\cite{ALSO:2012ECCV}, large scale classification and regression~\cite{HALSL:2011NIPS}, collaborative filtering~\cite{Zhou:2012KDD}, and recommendation~\cite{HeterogeneousH:2013KDD}. For indexing video sequences, a straightforward method is to independently compute binary codes for each key frames and use a set of hash code to represent video index. More recently, Ye et al. proposed a structure learning framework to derive a video hashing technique that incorporates both temporal and spatial structure information~\cite{Ye:2013ICCV}. In addition, advanced hashing methods are also developed for document search and retrieval. For instance, Wang et al. proposed to leverage both tag information and semantic topic modeling to achieve more accurate hash codes~\cite{SHTT:2013SIGIR}. Li et al. designed a two-stage unsupervised hashing framework for fast document retrieval~\cite{Twostage:2014ACL}. 

Hashing techniques have also been applied to the active learning framework to cope with big data applications. Without performing exhaustive test on all the data points, hyperplane hashing can help significantly speed up the interactive training sample selection procedure~\cite{Vijayanarasimhan:2013PAMI}\cite{Liu:2012ICML}\cite{Jain:2010NIPS}. In addition, a two-stage hashing scheme is developed to achieve fast query pair selection for large scale active learning to rank~\cite{Qian:2013ICDM}.

\section{Open Issues and Future Directions}
Despite the tremendous progress in developing a large array of hashing techniques, several major issues remain open. First, unlike the locality sensitive hashing family, most of the learning based hashing techniques lack the theoretical guarantees on the quality of returned neighbors. Although several recent techniques have presented theoretical analysis of the collision probability, they are mostly based on randomized hash functions~\cite{Jain:2010NIPS}\cite{Liu:2012ICML}\cite{Kulis:2009PAMI}. Hence, it is highly desired to further investigate such theoretical properties. Second, compact hash codes have been mostly studied for large scale retrieval problems. Due to their compact form, the hash codes also have great potential in many other large scale data modeling tasks such as efficient nonlinear kernel SVM classifiers~\cite{HSVM:cvpr2014} and rapid kernel approximation~\cite{Shi:2009JMLR}. A bigger question is: instead of using the original data, can one directly use compact codes to do generic unsupervised or supervised learning without affecting the accuracy? To achieve this, theoretically sound practical methods need to be devised. This will make efficient large-scale learning possible with limited resources, for instance on mobile devices. Third, most of the current hashing technicals are designed for given feature representations that tend to suffer from the semnatic gap. One of the possible future directions is to integrate representation learning with binary code learning using advanced learning schemes such as deep neural network. Finally, since heterogeneity has been an important characteristics of the big data applications, one of the future trends will be to design efficient hashing approaches that can leverage heterogeneous features and multi-modal data to improve the overall indexing quality. Along those lines, developing new hashing techniques for composite distance measures, i.e., those based on combinations of different distances acting on different types of features will be of great interest.


\label{sec:conclusion}

\bibliographystyle{IEEEtran}
\bibliography{Hashref}

\end{document}